\documentclass{article}
\usepackage{iclr2027_conference,times}

\usepackage{amsmath,amsfonts,bm}

\def\eqref#1{equation~\ref{#1}}

\def\1{\bm{1}}

\DeclareMathAlphabet{\mathsfit}{\encodingdefault}{\sfdefault}{m}{sl}
\SetMathAlphabet{\mathsfit}{bold}{\encodingdefault}{\sfdefault}{bx}{n}

\usepackage{amssymb}
\usepackage{graphicx}
\usepackage{etoolbox}
\usepackage{placeins}
\usepackage{capt-of}
\usepackage{booktabs}
\usepackage{longtable}
\usepackage{tabularx}
\usepackage{array}
\usepackage{makecell}
\usepackage{multirow}
\usepackage{siunitx}
\usepackage{colortbl}
\definecolor{tableaccent}{HTML}{EDF3F8}
\usepackage{hyperref}
\usepackage{url}

\makeatletter
\patchcmd{\@maketitle}{\vskip 0.3in minus 0.1in}{\vskip 4pt}{}
  {\PackageError{author-layout}{Could not adjust title spacing}{}}
\newcommand{\printappendixcontents}{%
  \phantomsection
  \pdfbookmark[0]{Appendix contents}{appendix-contents}
  \begingroup
  \hypersetup{hidelinks}
  \setcounter{tocdepth}{2}
  \setlength{\parskip}{0pt}
  \begin{center}\Large\bfseries Appendix\end{center}
  \vspace{0.5em}
  \noindent{\large\bfseries Table of Contents}\par
  \vspace{0.3em}\hrule\vspace{0.5em}
  \renewcommand{\l@subsection}{\@dottedtocline{2}{1.8em}{2.6em}}
  \@starttoc{apc}
  \vspace{0.7em}\hrule
  \endgroup
  \par\medskip
}
\newcommand{\enableappendixcontents}{%
  \let\paperaddcontentsline\addcontentsline
  \renewcommand{\addcontentsline}[3]{%
    \paperaddcontentsline{##1}{##2}{##3}
    \def\paperstream{##1}\def\papertoc{toc}%
    \ifx\paperstream\papertoc
      \addtocontents{apc}{\protect\contentsline{##2}{##3}{\thepage}{\@currentHref}\protected@file@percent}%
    \fi
  }%
}
\makeatother
\title{Traceable Human-to-Humanoid Sign \\Language Benchmarking}
\author{%
\textbf{Ao Liu$\quad$Shengeng Tang$^{*}$$\quad$Lechao Cheng$\quad$Yanbin Hao$\quad$Bingkun Bao$\quad$Richang Hong}\\
\normalfont Hefei University of Technology
{\normalfont\small\ttfamily (aoliu@mail.hfut.edu.cn, tangsg@hfut.edu.cn)}
}
\iclrfinalcopy
\hypersetup{
  pdfauthor={Ao Liu, Shengeng Tang, Lechao Cheng, Yanbin Hao, Bingkun Bao, Richang Hong}
}
\begin{document}
\maketitle
\pagestyle{plain}

\noindent\begin{minipage}{\linewidth}
\centering
\includegraphics[width=0.85\linewidth]{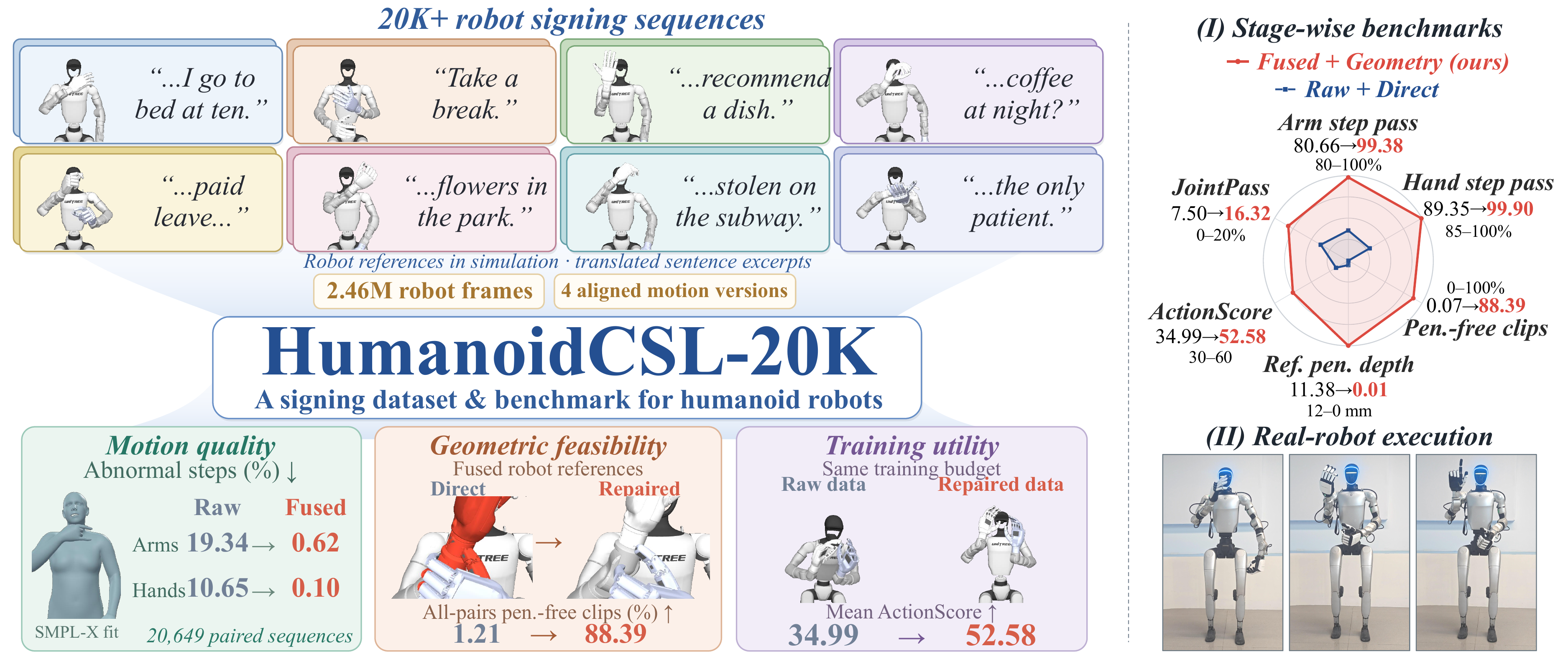}
\captionof{figure}{HumanoidCSL-20K links 20,648 signing sequences across four aligned versions of human and robot motion. Simulated examples and benchmark results show fewer abrupt joint changes and robot self-collisions after repair, plus higher control scores under the same training budget.}
\label{fig:overview}
\end{minipage}\par

\begin{abstract}
Sign data collection is costly, and teleoperation scales poorly, motivating reuse of large video corpora. Humanoid signing requires converting video-derived human motion into robot trajectories while preserving linguistic motion cues. Errors from fitting, human-motion repair, retargeting, robot geometry repair, and control are hard to separate from the final trajectory alone. We introduce HumanoidCSL-20K, a dataset and benchmark of 20,648 sentence-level Chinese Sign Language sequences, each with four aligned versions: the source, the repaired human motion, the direct robot reference, and the geometry-repaired robot reference. Observation-supported local human-motion repair, full-robot geometry repair, and cross-representation provenance make each transformation traceable. Paired evaluations measure human-motion continuity and content preservation, robot-reference feasibility, and physical execution. A sign-specific kinematic-reference protocol scores handshape, location, palm orientation, and inter-hand relation over the full planned motion. Full-corpus results show fewer abnormal arm / hand steps and less inter-hand and hand-body penetration after repair. Control experiments separate reference learnability from curriculum effects, while component scores expose remaining execution errors.
\end{abstract}

\begin{figure}[t]
    \centering
    \includegraphics[width=0.94\linewidth]{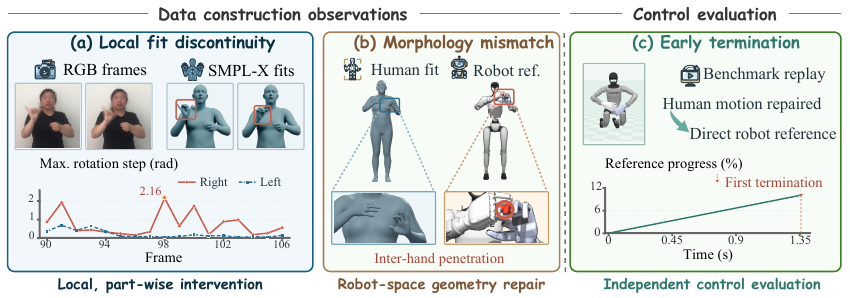}
\caption{Why fitted human motion is not directly robot-ready. (a)~Video fits contain local jumps and frozen hands. (b)~Plausible human motion penetrates the robot's own body after retargeting. (c)~A collision-free robot reference can still fail in execution. Replay stops at the first failure, and unreached frames score zero.}
    \label{fig:failures}
\end{figure}

\section{Introduction}
\label{sec:introduction}

Signing humanoid robots could enable face-to-face communication in public-service settings~\citep{qiao2025signbot,khan2026sign}. The meaning of a sign is carried by continuous and temporally structured motion, in which handshape, location, movement, palm orientation, and bimanual coordination collectively shape it. Reaching a correct final pose alone is therefore insufficient. A signing robot must follow a reference trajectory throughout the sign.

Motion capture provides accurate 3D signing data~\citep{jedlivcka2020sign,jedlivcka2022mc}, but every new sentence or signer requires a separate session with calibrated equipment, fluent signers, and post-processing. Finger motion is particularly costly to record: DexAvatar used nine cameras with data gloves and still had to filter implausible hand poses~\citep{kundu2026dexavatar}. In contrast, sign video corpora offer far broader linguistic coverage. CSL-Daily contains 20,654 continuous Chinese Sign Language videos from ten signers with translation and gloss annotations~\citep{zhou2021improving}, and SOKE provides SMPL-X fits for these videos~\citep{zuo2025signs}. These resources are a practical starting point for robot signing.

However, fitted human motion cannot drive a robot directly (Figure~\ref{fig:failures}). First, monocular fits are locally unreliable under fast articulation, occlusion, and blur; in the SOKE fits, 19.3\% of arm frames and 10.7\% of hand frames contain implausibly large rotation steps. Second, the robot's body differs from the signer's; direct retargeting of repaired motion still leaves 50.18\% of frames with inter-hand, hand-body, or intra-hand penetration. Third, a collision-free reference remains hard to execute, as the controller must balance the whole body while tracking twelve finger joints; a general-purpose AMP~\citep{peng2021amp} tracker stops within about ten frames in every replay.

These errors accumulate along the pipeline, making it difficult to trace a robot failure to its source when only the final motion is evaluated. Existing signing systems are evaluated end to end and provide no corpus-scale benchmark of aligned intermediate motions. We present HumanoidCSL-20K, a dataset and benchmark that retains four aligned versions of each sentence: the source fit, the repaired human motion, the direct robot reference, and the geometry-repaired robot reference. Each stage can thus be evaluated separately, and its effect isolated by replacing it while holding the other stages fixed. The main contributions of this paper are summarized as follows: 
\begingroup\setlength{\leftmargini}{1.2em}
\begin{itemize}
\setlength{\itemsep}{1pt}\setlength{\parsep}{0pt}
\item We present HumanoidCSL-20K, the first publicly released large-scale sign-language dataset for humanoid robots. It contains 20,648 sentence-level sequences, each with the source, the aligned motion, and the optimized humanoid data.
\item A local human-motion repair is developed, which cuts abnormal rotation steps to 0.62\% for arms and 0.10\% for hands, with no detectable recognition-BLEU drop against the source fits.
\item We introduce a geometry repair for robot references that reduces penetrating frames to 0.29\% and certifies 88.21\% of sequences as penetration-free, while changing finger joints by at most $20^\circ$.
\item We establish a stage-wise benchmark with execution metrics and a curriculum controller that learns body/wrist before finger tracking and completes every replay on repaired references.
\end{itemize}
\endgroup

\begin{figure}[h]
\centering
\includegraphics[width=\linewidth]{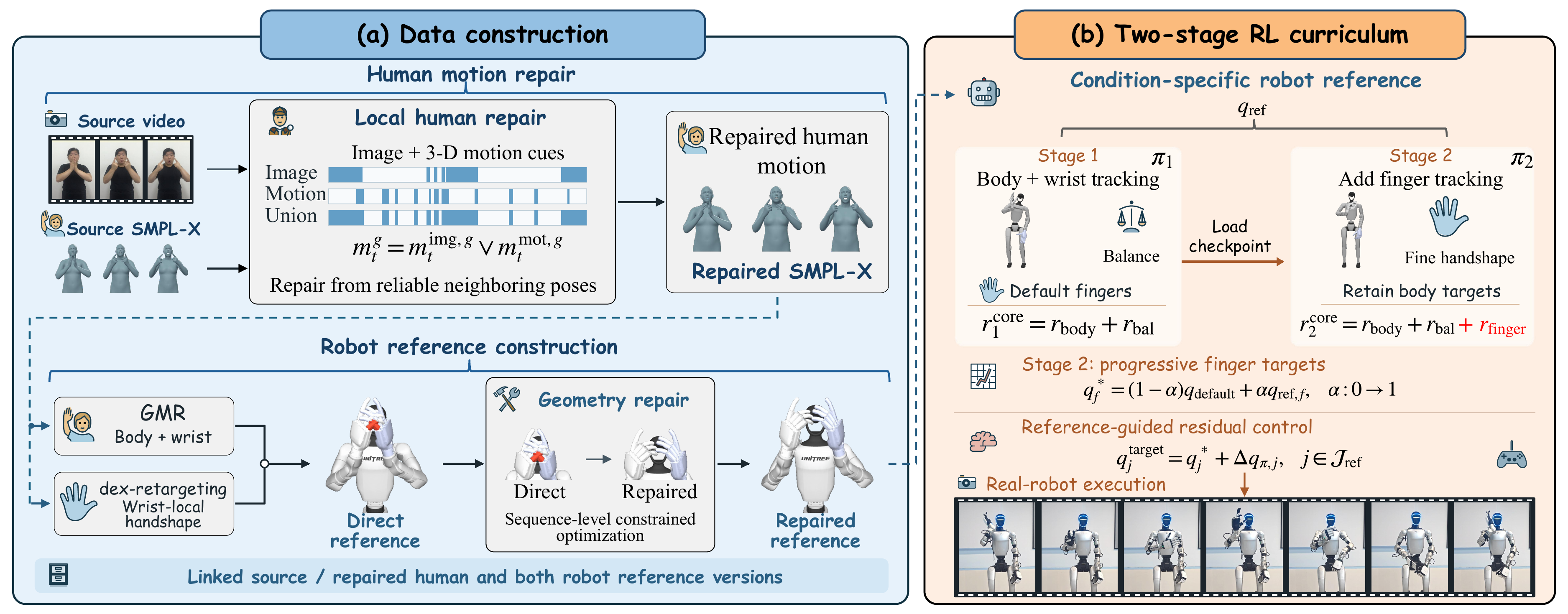}
\caption{Pipeline overview: (a) linked human and robot motion construction; (b) body-wrist tracking followed by fine handshape tracking.}
\label{fig:construction}
\end{figure}

\section{Related work}
\label{sec:related}

\noindent\textbf{Sign-language motion data.}\hspace{1em}3D signing data come from motion capture or from reconstruction of sign-language videos~\citep{yu2024signavatars}. Sign-language production methods generate human-level signing from glosses or text~\citep{tang2025gloss,zuo2025signs}, and reference-free metrics evaluate produced signing without ground-truth trajectories~\citep{cory2026backtranslation2}. These datasets and production methods stop at human motion. HumanoidCSL-20K extends the SOKE fits of CSL-Daily to robot references and records every repair applied to them; our metrics compare each manual component against a known reference trajectory.

\noindent\textbf{Human-to-robot motion transfer.}\hspace{1em}General Motion Retargeting maps body motion to humanoids~\citep{araujo2025retargeting}, DexPilot, AnyTeleop, and dex-retargeting map human hands to dexterous robot hands~\citep{handa2020dexpilot,qin2023anyteleop,qin2022one}, physics-informed retargeting scales across diverse embodiments~\citep{pan2025spider}, and kinodynamic retargeting enforces dynamic and contact feasibility~\citep{zhang2026kinodynamic,chen2026scalablewholebodymotiontransfer}. Combining separately retargeted body and hands can make the robot penetrate itself (Section~\ref{sec:geometry_results}), which our geometry repair removes over the whole sequence. For robot signing, production methods generate signing motion at the human level~\citep{tang2025signidd}, SignBot combines retargeting, learned control, and conversational interaction, and Khan et~al.\@ refine retargeted trajectories with vision-language guidance and collision mitigation. Both complete systems are evaluated end-to-end, whereas we release the intermediate motions for separate evaluation.

\noindent\textbf{Humanoid motion tracking.}\hspace{1em}Imitation and motion-prior methods~\citep{peng2018deepmimic,peng2022ase} and humanoid tracking systems~\citep{he2024learning,he2024omniho,cheng2024expressive,fu2024humanplus,liao2026beyondmimic} train policies to follow retargeted human motion, and constrained or residual formulations bound the deviation from the reference~\citep{yan2024ctrl,sun2026robotdancing}. Versatile controllers consolidate diverse control modes through motion imitation~\citep{he2025hover} or masked motion inpainting~\citep{tessler2024maskedmimic}, and universal motion representations provide reusable motor skills across large clip libraries~\citep{luo2024pulse}. Our controller adopts the per-reference tracking of BeyondMimic with residual actions and ramps in finger targets after body and wrist tracking is learned. Imperfect-demonstration methods estimate demonstration quality to reweight supervision~\citep{wu2019imitation,zhang2021confidence,beliaev2022imitation,cui2019uncertainty} or bridge discrete observations to continuous trajectories~\citep{tang2025discrete}; our repair records instead state how each segment was produced, so tracking errors can be grouped by the origin of each reference segment (Section~\ref{sec:control_results}).

\section{Dataset construction}
\label{sec:construction}

\subsection{Source motion and its errors}
\label{sec:sources}
\label{sec:source_diagnostics}

The source human motion $\mathcal{H}^{\mathrm{raw}}$ consists of the SOKE SMPL-X fits of CSL-Daily videos for 20,652 sequences. Local human-motion repair yields $\mathcal{H}^{\mathrm{fused}}$ and its direct robot reference $\mathcal{G}^{\mathrm{direct}}$ for 20,649 sequences, and geometry repair yields $\mathcal{G}^{\mathrm{repair}}$ for 20,648 (Figure~\ref{fig:construction}). All versions share sequence identifiers, source-video timestamps, and a human-to-robot body-part mapping, so every frame traces back to the video and its annotation (Appendix~\ref{app:data}).

\begin{figure}[!htbp]
\centering
\includegraphics[width=0.8\linewidth]{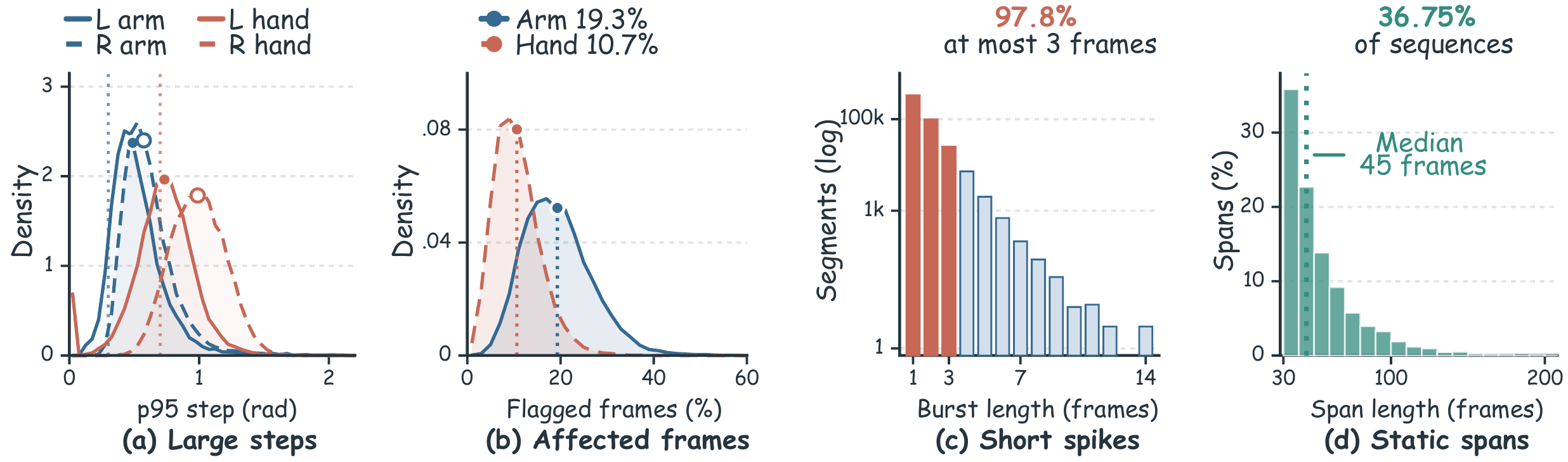}
\caption{Source-motion diagnostics over 20,652 sequences: (a) rotation-step distributions, (b) affected-frame rates, (c) short motion bursts, and (d) long static-hand spans.}
\label{fig:quality_distributions}
\end{figure}

Figure~\ref{fig:quality_distributions} shows two error patterns in $\mathcal{H}^{\mathrm{raw}}$. On average, 19.3\% of arm frames and 10.7\% of hand frames contain a rotation step above 0.30 and 0.70\,rad, respectively (panels a, b). Most of these errors are brief, as 97.8\% of candidate segments last at most three frames (panel c). The second pattern is long static hand spans, which last 45 frames at the median and occur in 36.75\% of sequences (panel d). Brief jumps can be bridged from neighboring frames, whereas long spans need a held pose. Both patterns are local in time and body part, so the repair is local rather than sequence-wide.

\subsection{Local human-motion repair}
\label{sec:human_repair}

We repair the left and right arms and hands as four separate part groups. A frame is flagged for a group when either of two cues fires. The image cue uses RTMPose-l WholeBody keypoints~\citep{jiang2023rtmpose} and marks low keypoint confidence or visibility, blur, and abrupt keypoint or bounding-box changes. The motion cue marks large fitted rotation steps. For part group $g$ with joints $\mathcal{J}_g$ and threshold $\theta_g$ (0.30\,rad for arms, 0.70\,rad for hands), the step cue and the repair mask are
\begin{equation}
\label{eq:repair_mask}
m^{\mathrm{mot},g}_t=\mathbf{1}\Bigl[\max_{j\in\mathcal{J}_g}d_{SO(3)}\bigl(R_{t-1,j},R_{t,j}\bigr)>\theta_g\Bigr],\qquad
m^g_t=m^{\mathrm{img},g}_t\lor m^{\mathrm{mot},g}_t,
\end{equation}
where $m^{\mathrm{img},g}_t$ is the image cue; the motion cue also flags short oscillations (Appendix~\ref{app:human}). We call the repaired motion Fused because its mask is the union of the two cues. Each flagged segment is rebuilt from reliable neighboring frames. A short gap between reliable anchor frames $a<b$ is filled by spherical interpolation,
\begin{equation}
\label{eq:slerp_fill}
R_{t,j}=\operatorname{Slerp}\Bigl(R_{a,j},R_{b,j},\tfrac{t-a}{b-a}\Bigr),\qquad a<t<b,
\end{equation}
and a longer gap holds $R_{a,j}$ and then blends into the next reliable observation. Remaining flagged arm windows receive light smoothing, while wrists and fingers are not smoothed. For every frame and part group, the output records which operation was applied, with a fixed observation-support score for how directly that operation is supported by the video; shared timestamps carry these provenance records to the robot references and policy rollouts (Appendices~\ref{app:human} and~\ref{app:provenance}).

The abnormal-step metric in Section~\ref{sec:benchmark} uses the same 0.30 and 0.70\,rad thresholds as the motion cue, so for Fused it measures how completely the detected jumps are removed. We therefore also report two results that do not depend on these thresholds: a frozen sign recognizer shows no BLEU loss after repair (Section~\ref{sec:human_results}), and single-stage policies trained on Fused references complete 58 of 60 replays, compared with 34 of 60 for Raw references under the same budget (Section~\ref{sec:control_results}).

\subsection{Robot-reference construction and geometry repair}
\label{sec:geometry_repair}

General Motion Retargeting maps $\mathcal{H}^{\mathrm{fused}}$ to the G1 body and wrist poses, and dex-retargeting maps each human hand to an Inspire hand. Their combination is the direct reference $\mathcal{G}^{\mathrm{direct}}$. The body and hands are retargeted separately, so neither solver checks the assembled robot for collisions. Since the robot's links are also bulkier than human limbs, signing poses that bring the hands close to each other or to the body become penetrations. Geometry repair removes these penetrations in two steps: an arm stage followed by a finger stage.

The arm stage optimizes the 14 shoulder, elbow, and wrist joints $\boldsymbol{x}_t$ over the whole sequence, with the root, waist, legs, and fingers fixed:
\begin{equation}
\label{eq:geometry_repair}
\begin{aligned}
\min_{\boldsymbol{x}_{1:N}}\ &\sum_{t=1}^{N}\Bigl(\mathcal{E}_{\mathrm{track}}(\boldsymbol{x}_t)+\lambda_r\,\mathcal{E}_{\mathrm{rel}}(\boldsymbol{x}_t)\Bigr)+\mathcal{E}_{\mathrm{smooth}}(\boldsymbol{x}_{1:N})\\
\text{s.t.}\ &d_k\bigl(\boldsymbol{x}(\tau)\bigr)\geq\gamma,\quad k\in\mathcal{K}_{\mathrm{body}},\ \tau\in\Omega_M.
\end{aligned}
\end{equation}
Here $\mathcal{E}_{\mathrm{track}}$ keeps the shoulder, elbow, and wrist frames $b\in\mathcal{B}$ close to their scaled human targets $T^{*}_{b,t}$, and $\mathcal{E}_{\mathrm{smooth}}$ penalizes temporal differences:
\begin{equation}
\label{eq:arm_terms}
\begin{aligned}
\mathcal{E}_{\mathrm{track}}(\boldsymbol{x}_t)&=\sum_{b\in\mathcal{B}}\bigl\lVert W_b\operatorname{Log}\bigl(T_b(\boldsymbol{x}_t)^{-1}T^{*}_{b,t}\bigr)^{\vee}\bigr\rVert_2^2,\\
\mathcal{E}_{\mathrm{smooth}}(\boldsymbol{x})&=\lambda_v\lVert D_1\boldsymbol{x}\rVert_2^2+\lambda_a\lVert D_2\boldsymbol{x}\rVert_2^2+\lambda_j\lVert D_3\boldsymbol{x}\rVert_2^2,
\end{aligned}
\end{equation}
where $\operatorname{Log}(\cdot)^{\vee}$ maps a relative pose to its six-dimensional twist, $W_b$ weights its position and rotation parts, and $D_d$ is the $d$-th temporal difference (Appendix~\ref{app:solver}). The inter-hand relation term
\begin{equation}
\label{eq:bilateral_relation_main}
\mathcal{E}_{\mathrm{rel}}(\boldsymbol{x}_t)=\bigl\lVert(\boldsymbol{p}_{R,t}-\boldsymbol{p}_{L,t})-(\boldsymbol{p}^h_{R,t}-\boldsymbol{p}^h_{L,t})\bigr\rVert_2^2
\end{equation}
keeps the vector between the robot wrists $\boldsymbol{p}_{L,t},\boldsymbol{p}_{R,t}$ close to that between the scaled human wrists $\boldsymbol{p}^h_{L,t},\boldsymbol{p}^h_{R,t}$. Every inter-hand and hand-body pair $k\in\mathcal{K}_{\mathrm{body}}$ must keep a clearance $\gamma=3$\,mm, at frames and at $M=8$ instants between consecutive frames ($\Omega_M$), so the arms cannot cross between frames. Joint, speed, acceleration, and jerk limits also apply. It is solved by sequential quadratic programming with linearized distances and penalized slack (Appendix~\ref{app:solver}).

The finger stage fixes the arms and adjusts the 12 active finger joints, six per hand, with passive joints following the mechanical coupling. With $\Delta\boldsymbol{q}^f_t$ the change from the direct handshape and $\mathcal{K}_f$ the pairs involving a finger segment, it solves
\begin{equation}
\label{eq:finger_repair}
\begin{aligned}
\min_{\Delta\boldsymbol{q}^f,\boldsymbol{s}}\ &\sum_{t=1}^{N}\lVert\Delta\boldsymbol{q}^f_t\rVert_2^2+\mathcal{E}_{\mathrm{smooth}}(\Delta\boldsymbol{q}^f)+\rho^2\lVert\boldsymbol{s}\rVert_2^2\\
\text{s.t.}\ &|\Delta q^f_{t,j}|\leq\theta_{\max},\quad d_k\bigl(\boldsymbol{q}(\tau)\bigr)+s_{k,\tau}\geq\gamma_f,\quad s_{k,\tau}\geq0,\quad k\in\mathcal{K}_f,\ \tau\in\Omega_M,
\end{aligned}
\end{equation}
where $\theta_{\max}=20^\circ$ is a hard budget and the clearance $\gamma_f=0.2$\,mm is enforced through the penalized slack $\boldsymbol{s}$; where the two conflict, the handshape is kept and slight contact remains.

Geometry repair produces $\mathcal{G}^{\mathrm{repair}}$ for 20,648 sequences that pass joint-range, speed, finger-coupling, and numerical checks. Let $\delta_{n,u}=[-\min_{k\in\mathcal{K}_{\mathrm{eval}}}d_k(\boldsymbol{q}_{n,u})]_+$ be the worst-pair penetration depth of sequence $n$ at sample $u$ over all audited pairs $\mathcal{K}_{\mathrm{eval}}=\mathcal{K}_{\mathrm{body}}\cup\mathcal{K}_{\mathrm{intra}}$, including intra-hand pairs. A sequence is \emph{certified} when
\begin{equation}
\label{eq:certified}
c_n=\prod_{u\in\mathcal{U}^{30}_n\cup\,\mathcal{U}^{50}_n}\mathbf{1}\bigl[\delta_{n,u}\leq\epsilon\bigr]=1,\qquad \epsilon=1\,\mu\mathrm{m},
\end{equation}
where $\mathcal{U}^{30}_n$ contains the 30\,Hz frames with 16 interpolated instants per interval and $\mathcal{U}^{50}_n$ the 50\,Hz control grid with three interior samples per interval. The 18,214 certified sequences (88.21\%) are penetration-free; the other 2,434 retain contact on 2.2\% of their samples, classified in Appendix~\ref{app:intrahand_results}. All statistics use the full $\mathcal{G}^{\mathrm{repair}}$, and the released flag lets users select the certified subset.

\section{Benchmark tasks and evaluation}
\label{sec:benchmark}

\subsection{Stage-wise benchmark tasks}
\label{sec:tasks}

The benchmark defines one task per transformation in Figure~\ref{fig:construction}. Each task fixes its input, output, and evaluation set, so a new method can replace one step and be compared with the released versions.

\paragraph{Human-motion repair.}
The input is $\mathcal{H}^{\mathrm{raw}}$ and the output is a repaired human motion. The released baselines are Raw, Global smoothing (Gaussian smoothing of the whole sequence; Appendix~\ref{app:human}), and Fused. Continuity is measured by the abnormal-step rate of each part group $g$, the share of steps whose largest within-group rotation exceeds $\theta_g$ (Equation~\ref{eq:repair_mask}),
\begin{equation}
\label{eq:abnormal_rate}
A^g=\frac{100}{N}\sum_{n=1}^{N}\frac{1}{2|\mathcal{T}_n|}\sum_{s\in\{L,R\}}\sum_{t\in\mathcal{T}_n}\mathbf{1}\Bigl[\max_{j\in\mathcal{J}_{g,s}}d_{SO(3)}\bigl(R_{t-1,j},R_{t,j}\bigr)>\theta_g\Bigr],
\end{equation}
where $s$ is the body side, $\mathcal{T}_n$ the steps of sequence $n$, and $N=20{,}649$. At 30\,Hz these steps correspond to 9 and 21\,rad/s, far above the arm (4--5.4\,rad/s) and finger (4.8\,rad/s) speed limits of the robot references (Appendix~\ref{app:solver}), so a remaining abnormal step cannot be executed as is. Because the motion cue of Fused uses the same thresholds (Section~\ref{sec:human_repair}), content preservation is measured separately: a frozen pose-only Uni-Sign recognizer~\citep{li2025uni} translates each version of 75 held-out sequences (38 development and 37 test examples not used in its training), scored by BLEU.

\paragraph{Robot-reference construction.}
The input is a human motion and the output is a reference for the G1 robot with Inspire hands. The released baselines are direct retargeting of Raw, Global, and Fused motion, and $\mathcal{G}^{\mathrm{repair}}$. Feasibility is audited over all contact pairs of the assembled robot, $\mathcal{K}_{\mathrm{eval}}$: inter-hand, hand-body, and intra-hand pairs. Each reference is evaluated on the 50\,Hz grid samples $\mathcal{U}^{\mathrm{grid}}_n$ and the three interior samples per interval $\mathcal{U}^{\mathrm{int}}_n$, whose union is $\mathcal{U}^{50}_n$. With the depth $\delta_{n,u}$ of Equation~\ref{eq:certified}, the penetration rates and the penetration-free share are
\begin{equation}
\label{eq:geometry_metrics_main}
P^{\star}_{\mathrm{pen}}=\frac{100}{N}\sum_{n=1}^{N}\frac{1}{|\mathcal{U}^{\star}_n|}\sum_{u\in\mathcal{U}^{\star}_n}\mathbf{1}[\delta_{n,u}>\epsilon],\qquad
P_{\mathrm{free}}=\frac{100}{N}\sum_{n=1}^{N}\prod_{u\in\mathcal{U}^{50}_n}\mathbf{1}[\delta_{n,u}\leq\epsilon],
\end{equation}
where $\star\in\{\mathrm{grid},\mathrm{int}\}$ and $N=20{,}648$; the mean depth averages $\delta_{n,u}$ in the same way. Computing each quantity per sequence keeps long clips from dominating (Appendix~\ref{app:geometry}). Certification (Section~\ref{sec:geometry_repair}) also checks the 30\,Hz grid and is therefore stricter than being penetration-free. Two further measures show how much repair changes the motion: the RMS change of the right-minus-left wrist vector and the RMS change of the active finger joints.

\paragraph{Robot execution.}
The input is a robot reference and the output is a simulated controller rollout that tracks the whole reference with fixed timestamps. The evaluation set is 20 randomly sampled sentence-level motions, each trained with three seeds (Appendix~\ref{app:production}). Rollouts are scored by the sign-specific metrics below and by penetration over the same pair set $\mathcal{K}_{\mathrm{eval}}$ as the reference audit.

\subsection{Sign-specific execution metrics}
\label{sec:execution_score}
\begingroup
\setlength{\abovedisplayskip}{3.5pt plus 1pt minus 1pt}
\setlength{\belowdisplayskip}{3.5pt plus 1pt minus 1pt}

A whole-body joint average can hide the few errors that decide a sign, such as a wrong handshape or crossed wrists, so we score the manual components separately. Let $B$ be a fixed torso link and $W\in\{L,R\}$ a hand. Its torso-relative wrist position and palm orientation are $\widetilde{\boldsymbol{p}}_W=R_B^\top(\boldsymbol{p}_W-\boldsymbol{p}_B)$ and $\widetilde R_W=R_B^\top R_W$. At reference frame $t$, the executed and reference states give four errors for handshape, hand location, palm orientation, and inter-hand relation:
\begin{equation}
\label{eq:sign_errors}
\begin{aligned}
e_t^{\mathrm{hand}}&=\max_{j\in\mathcal{J}_{\mathrm{hand}}}|q^{\mathrm{exec}}_{t,j}-q^{\mathrm{ref}}_{t,j}|,\quad &
e_t^{\mathrm{pos}}&=\max_{W\in\{L,R\}}\lVert\widetilde{\boldsymbol{p}}^{\mathrm{exec}}_{t,W}-\widetilde{\boldsymbol{p}}^{\mathrm{ref}}_{t,W}\rVert_2,\\
e_t^{\mathrm{ori}}&=\max_{W\in\{L,R\}}d_{SO(3)}(\widetilde R^{\mathrm{exec}}_{t,W},\widetilde R^{\mathrm{ref}}_{t,W}),\quad &
e_t^{\mathrm{rel}}&=\lVert(\widetilde{\boldsymbol{p}}_R-\widetilde{\boldsymbol{p}}_L)^{\mathrm{exec}}_t-(\widetilde{\boldsymbol{p}}_R-\widetilde{\boldsymbol{p}}_L)^{\mathrm{ref}}_t\rVert_2.
\end{aligned}
\end{equation}
Here $\mathcal{J}_{\mathrm{hand}}$ contains the twelve active finger joints, and the maximum keeps a local failure from being averaged away. Movement and timing, the remaining cues of a sign, enter through frame-by-frame comparison without temporal warping, which penalizes delayed or desynchronized hands.

Each component $x\in\{\mathrm{hand},\mathrm{pos},\mathrm{ori},\mathrm{rel}\}$ has a nominal scale $\sigma_x$ of 0.30\,rad, 0.10\,m, 0.50\,rad, and 0.10\,m, which also serves as its pass threshold. For handshape, 0.30\,rad is about one fifth of the 1.44\,rad flexion range of the four Inspire fingers; for palm orientation, 0.50\,rad ($29^\circ$) is below the $45^\circ$ midpoint between canonical palm directions; for location and relation, 0.10\,m is a declared tolerance, coarser than the spacing of some facial locations (Appendix~\ref{app:articulation_spacing}). Scaling all thresholds together changes some rankings, but Fused + Geometry stays above the other references (Appendix~\ref{app:threshold_auc}).

ActionScore maps each error to a fidelity $\psi_x(e)=2^{-(e/\sigma_x)^2}$, which is 1 at zero error and 0.5 at one nominal scale, and averages it over the planned motion:
\begin{equation}
\label{eq:action_score}
C^x_{i,s}=\frac{100}{T_i}\sum_{t=0}^{T_i-1}A_{i,s,t}\,\psi_x(e_t^x),\qquad
\operatorname{AS}_{i,s}=\frac{1}{4}\sum_x C^x_{i,s}.
\end{equation}
Here $T_i$ is the planned length of motion $i$, and $A_{i,s,t}=1$ if frame $t$ is reached before failure under seed $s$, so unreached frames score zero. The four components are weighted equally, and a score of 100 requires exact execution throughout. JointPass is the share of planned frames that are reached, pass all four thresholds, and satisfy the physical constraints:
\begin{equation}
\label{eq:joint_pass}
\operatorname{JointPass}_{i,s}=\frac{100}{T_i}\sum_{t=0}^{T_i-1}
A_{i,s,t}\,G_{i,s,t}\prod_{x}\mathbf{1}[e_t^x\leq\sigma_x],
\end{equation}
where $G_{i,s,t}=1$ if every pair in $\mathcal{K}_{\mathrm{eval}}$ penetrates by at most $\epsilon$ and every joint stays within its range up to $10^{-3}$\,rad. ActionScore credits partial accuracy continuously, whereas JointPass requires all components and constraints at once. Per-component pass rates are reported in Appendix~\ref{app:control_metrics}.
\endgroup

\subsection{Replay protocol}
\label{sec:protocol}

Two replay regimes answer different questions. In self-target replay, each policy is trained and evaluated on its own processed reference, which measures how learnable that reference is. In fixed-reference replay, all controllers track the same $\mathcal{G}^{\mathrm{repair}}$ trajectory, which isolates controller differences such as the curriculum. Paired conditions share the robot model, initial state, timestamps, evaluation grid, and failure rules. Each replay tracks the full finger target from the first frame and stops at the first fall, task termination, or numerical failure. Unreached frames stay in all denominators, so stopping early cannot avoid difficult segments. Tracking errors on reached frames are reported together with coverage, and penetration uses every recorded state. For a metric $\mathcal{M}$ and two conditions $A$ and $B$ evaluated on motions $\mathcal{I}$ and seeds $\mathcal{S}$, the reported paired difference averages seeds within each motion and weights motions equally,
\begin{equation}
\label{eq:paired_diff}
\bar{\Delta}_{A,B}=\frac{1}{|\mathcal{I}|}\sum_{i\in\mathcal{I}}\frac{1}{|\mathcal{S}|}\sum_{s\in\mathcal{S}}\bigl(\mathcal{M}^{A}_{i,s}-\mathcal{M}^{B}_{i,s}\bigr),
\end{equation}
and its 95\% interval resamples motions (Appendices~\ref{app:control_metrics} and~\ref{app:statistics}).

\section{Experiments}
\label{sec:experiments}

\subsection{Experimental setup}
\label{sec:setup}

\paragraph{Controller.}
Our controller follows the per-reference tracking of BeyondMimic: a PPO policy is trained for each reference at 50\,Hz, with 200\,Hz physics, and outputs residuals around the reference joint targets. Signing couples whole-body balance with fine finger motion, so the curriculum first learns balance and wrist placement, which carry hand location and palm orientation, and adds handshape afterwards. The first 1,500 of 3,000 iterations track the body and wrists with the fingers at a default pose. The remaining 1,500 add finger tracking, and a ramp moves the finger targets from the default pose to the reference. The no-ramp variant applies finger targets at once, and the single-stage variant tracks everything for all 3,000 iterations. All variants share the network, PPO settings, and interaction budget (Appendix~\ref{app:control} lists all settings).

\paragraph{Evaluation.}
Human-motion and geometry results use the full corpus (Section~\ref{sec:tasks}). Control results use 20 randomly sampled motions with seeds 42, 1234, and 777, giving 60 replays per condition; paired differences and their 95\% intervals follow Equation~\ref{eq:paired_diff}. As an external baseline, AMP is trained natively as one ProtoMotions policy over all 20 motions with the same interaction budget and uses the same scoring (Appendix~\ref{app:control}).

\subsection{Human-motion repair}
\label{sec:human_results}

\begin{figure}[!htbp]
\centering
\includegraphics[width=\linewidth]{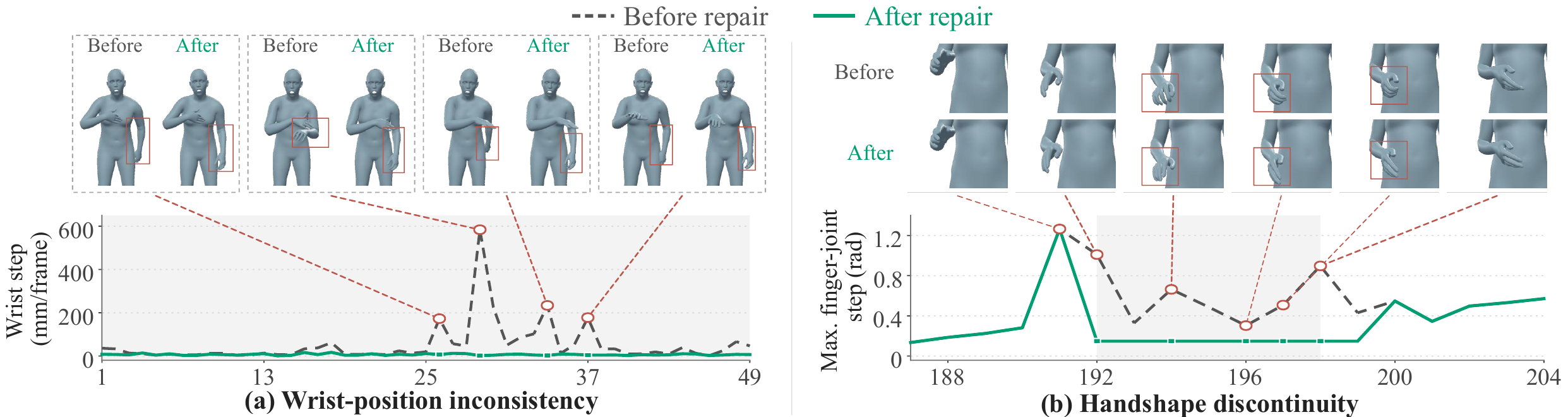}
\caption{Representative local repair of isolated (a) wrist and (b) handshape discontinuities while retaining neighboring anchors. Boxes mark corrections, and leaders connect poses to curves.}
\label{fig:human_repair_main}
\end{figure}

\newcommand{\BoundaryEligible}{20,636}
\newcommand{\BoundaryExcluded}{13}
\newcommand{\BoundaryEdges}{646,083}
Table~\ref{tab:human} shows that Global smoothing and Fused both remove most abnormal steps but differ in what they preserve. Global changes every frame and lowers BLEU-1 and BLEU-2 relative to Raw. Fused edits only flagged segments: outside the repair mask it leaves every hand pose unchanged and moves arm joints by $2.4^\circ$ on average (Appendix~\ref{app:human_metrics}). Its BLEU stays within the 95\% interval of Raw for every order (BLEU-4: $-0.29$, $[-2.01,1.34]$), although the recognizer was trained on raw fits and would penalize any content removed by repair. Fused exceeds Global by 5.45 BLEU-1 and 3.84 BLEU-2 points (BLEU-2 interval $[0.97,6.80]$), and both gains hold under a Bonferroni adjustment over the four BLEU orders. In Figure~\ref{fig:human_repair_main}(b), each frame looks plausible, but the handshape closes and reopens across frames 196--198; Fused replaces the event and keeps the anchor at frame 191.

\subsection{Robot-reference construction}
\label{sec:geometry_results}

\begin{figure}[t]
\begingroup
\setlength{\abovecaptionskip}{2pt}
\setlength{\belowcaptionskip}{2pt}
\centering
\begin{minipage}[t]{0.485\linewidth}
\centering
\captionof{table}{Human-motion continuity ($n=20{,}649$) and content preservation ($n=75$).}
\label{tab:human}
\scriptsize
\setlength{\tabcolsep}{2.2pt}
\renewcommand{\arraystretch}{1.08}
\begin{tabular}{@{}lcccc@{}}
\toprule
Method & \makecell{Arm abn.\\(\%) $\downarrow$} & \makecell{Hand abn.\\(\%) $\downarrow$} & \makecell{BLEU-2\\$\uparrow$} & \makecell{BLEU-4\\$\uparrow$} \\
\midrule
Raw & 19.34 & 10.65 & \textbf{41.46} & \textbf{26.59} \\[2.9pt]
Global smoothing & 1.76 & 0.24 & 37.33 & 24.50 \\[2.9pt]
\rowcolor{tableaccent} Fused & \textbf{0.62} & \textbf{0.10} & 41.17 & 26.29 \\[2.9pt]
\bottomrule
\end{tabular}
\end{minipage}
\hfill
\begin{minipage}[t]{0.485\linewidth}
\centering
\captionof{table}{Robot-reference geometry over all audited pairs at 50\,Hz ($n=20{,}648$).}
\label{tab:geometry}
\scriptsize
\setlength{\tabcolsep}{2.2pt}
\renewcommand{\arraystretch}{1.08}
\begin{tabular}{@{}lcccc@{}}
\toprule
Method & \makecell{Pen.\\(\%) $\downarrow$} & \makecell{Inter-fr.\\(\%) $\downarrow$} & \makecell{Depth\\(mm) $\downarrow$} & \makecell{Pen.-free\\(\%) $\uparrow$} \\
\midrule
Raw + Direct & 53.11 & 53.16 & 11.38 & 0.07 \\
Global + Direct & 52.47 & 52.50 & 11.00 & 0.21 \\
Fused + Direct & 50.18 & 50.20 & 8.28 & 1.21 \\
\rowcolor{tableaccent} Fused + Geometry ($\mathcal{G}^{\mathrm{repair}}$) & \textbf{0.29} & \textbf{0.30} & \textbf{0.01} & \textbf{88.39} \\
\bottomrule
\end{tabular}
\end{minipage}
\par\vspace{10pt}
\includegraphics[width=\linewidth]{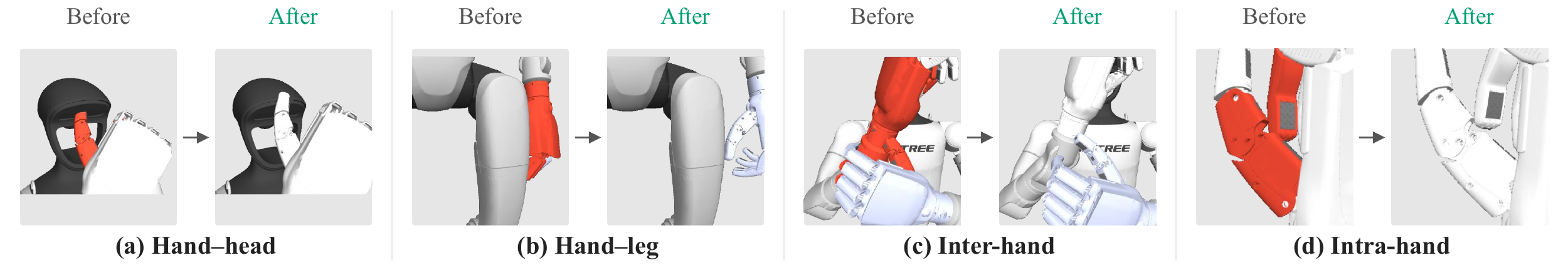}
\caption{Geometry repair examples for hand-head, hand-leg, inter-hand, and intra-hand penetration; red marks collisions.}
\label{fig:geometry_repair_main}
\endgroup
\end{figure}

Table~\ref{tab:geometry} shows that human-side processing alone does not make references feasible. With direct retargeting, about half of all frames penetrate for every human input, and at most 1.21\% of sequences are penetration-free. Geometry repair (Figure~\ref{fig:geometry_repair_main}) lowers penetrating frames from 50.18\% to 0.29\% and mean depth from 8.28 to 0.01\,mm, and it certifies 88.21\% of sequences (Section~\ref{sec:geometry_repair}). Interior samples follow the grid, so collisions are removed rather than moved between frames.

The changes are concentrated where the robot's body requires them. Its forearm and wrist links have a 4--5\,cm collision radius, so forearm crossings that nearly touch on a human signer must be opened on the robot. The repair widens the inter-wrist vector by 7.96\,cm RMS for this purpose, while the inter-hand relation term in Equation~\ref{eq:geometry_repair} keeps the rest of the relative hand placement close to the human reference. Handshapes change by only $1.40^\circ$ RMS over the active finger joints (Appendix~\ref{app:intrahand_results}).

\subsection{Robot execution}
\label{sec:control_results}

\begin{table}[t]
\centering
\caption{Execution on 20 motions and three seeds. Reading down, reference repair improves both of our controllers; reading across, our controllers far exceed AMP, and the full curriculum is best on the repaired reference.}
\label{tab:control_main}
\footnotesize
\setlength{\tabcolsep}{3pt}
\renewcommand{\arraystretch}{1.1}
\begin{tabular*}{\linewidth}{@{\extracolsep{\fill}}lccccccccc@{}}
\toprule
& \multicolumn{3}{c}{AMP} & \multicolumn{3}{c}{Ours, single-stage} & \multicolumn{3}{c}{Ours, two-stage + ramp} \\
\cmidrule(lr){2-4}\cmidrule(lr){5-7}\cmidrule(lr){8-10}
Reference & AS $\uparrow$ & Fin.\ $\uparrow$ & Pen.\ $\downarrow$ & AS $\uparrow$ & Fin.\ $\uparrow$ & Pen.\ $\downarrow$ & AS $\uparrow$ & Fin.\ $\uparrow$ & Pen.\ $\downarrow$ \\
\midrule
Raw + Direct & 2.73 & 0.0 & 32.24 & 34.99 & 56.7 & 36.15 & 35.51 & 60.0 & 35.53 \\
Fused + Direct & 2.68 & 0.0 & 26.39 & 49.15 & 96.7 & 25.97 & 49.29 & 98.3 & 26.08 \\
Fused + Geometry & 2.79 & 0.0 & 23.50 & 52.58 & 100.0 & 16.09 & \cellcolor{tableaccent}\textbf{54.10} & \cellcolor{tableaccent}\textbf{100.0} & \cellcolor{tableaccent}\textbf{13.03} \\
\bottomrule
\end{tabular*}
\par\smallskip
\begin{minipage}{\linewidth}\footnotesize
AS: ActionScore; Fin.: finished replays (\%); Pen.: penetrating rollout frames (\%) over all audited pairs, including intra-hand pairs. Unreached frames score zero. Without the ramp, two-stage training on Fused + Geometry reaches 53.44 / 100.0 / 13.55. Table~\ref{tab:control} in Appendix~\ref{app:seeds} adds JointPass, depth, and Global smoothing.
\end{minipage}
\end{table}

\paragraph{Reference processing.}
Down each column group of Table~\ref{tab:control_main}, every policy is trained on its own reference (self-target replay, Section~\ref{sec:protocol}). Repairing the source motion makes references much easier to learn: with single-stage training, Fused raises completion from 56.7\% to 96.7\% and ActionScore from 34.99 to 49.15 (paired gain 14.16, 95\% interval $[7.23,21.43]$). Geometry repair adds a further gain on top of Fused in both curricula. It raises ActionScore to 52.58 and 54.10 and JointPass to 16.32\% and 17.98\%, and under the full curriculum it halves rollout penetration from 26.08\% to 13.03\% while reducing depth from 0.663 to 0.112\,mm (Table~\ref{tab:control}). Rollouts still penetrate more often than the reference (13.03\% versus 0.29\%), so execution is evaluated as a separate task.

\paragraph{Curriculum.}
The last row of Table~\ref{tab:control_main} fixes the Fused + Geometry reference. All our curricula complete every replay, and the full curriculum gives the highest ActionScore and JointPass. Its ActionScore exceeds single-stage and no-ramp training by 1.52 and 0.66 points and is higher for every training seed. The clearest gains are in hand placement: wrist RMS falls from 13.97 to 13.42\,cm (95\% paired interval $[-1.04,-0.04]$) and palm RMS from $40.97^\circ$ to $39.19^\circ$ ($[-3.24,-0.30]$). Per-component pass rates also favor the full curriculum in the three-seed mean (Appendix~\ref{app:control_metrics}).

\paragraph{General-purpose tracker.}
AMP stops within about ten frames on every reference (Table~\ref{tab:control_main}), during the transition from the initial posture, and its ActionScore stays below 3. Repairing the reference does not change this outcome, whereas our controller completes every replay on the repaired reference. Under this protocol, a general-purpose motion prior does not transfer directly to dexterous signing, whereas reference-conditioned tracking designed for signing executes the full motion.

\paragraph{Using the repair records.}
Because the repair records follow each reference into its rollouts, execution errors can be grouped by how each reference segment was produced. Right-arm tracking error is 0.162\,rad RMS on observed frames and 0.240\,rad on long held spans (Appendix~\ref{app:provenance_stratified}), so a controller developer can separate errors on held segments from errors on observed motion. The records can also guide training: weighting supervision by the records gives lower high-support finger RMS than removing low-support motions (0.059 versus 0.066\,rad).

\begin{figure}[!htbp]
\centering
\setlength{\abovecaptionskip}{2pt}
\setlength{\belowcaptionskip}{2pt}
\includegraphics[width=\linewidth]{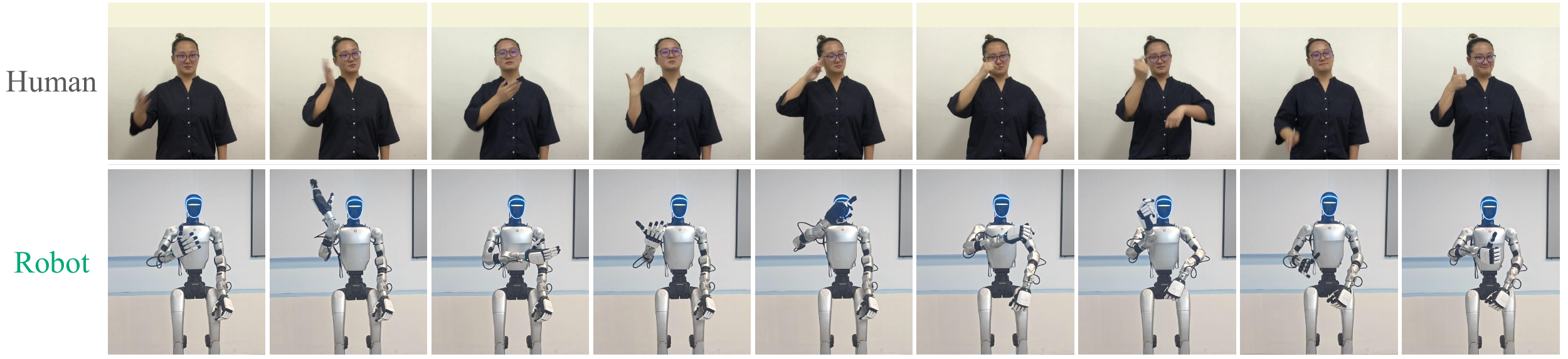}
\caption{Replay on a physical Unitree G1 humanoid: source-signer frames (top) and the corresponding robot replay (bottom).}
\label{fig:real_robot_main}
\end{figure}

\paragraph{Hardware replay.}
Figure~\ref{fig:real_robot_main} shows a physical Unitree G1 with two Inspire hands replaying a signing sequence alongside the source video frames; Appendix~\ref{app:real} describes the setup and replays.

\section{Conclusion and limitations}
\label{sec:conclusion}

HumanoidCSL-20K provides 20,648 sentence-level Chinese Sign Language sequences, each with four aligned versions: the source, the repaired human motion, the direct robot reference, and the geometry-repaired robot reference. Observation-supported local human-motion repair removes nearly all abnormal steps without a detectable loss in frozen-recognizer BLEU. Full-robot geometry repair lowers penetrating frames from 50.18\% to 0.29\% and certifies 88.21\% of sequences, while changing handshapes by only $1.40^\circ$ RMS. In execution, repaired references are easier to learn and collide less in rollouts, and our curriculum controller completes every replay on them. Because every version is kept and aligned, each step of the pipeline can be evaluated and replaced on its own.

\paragraph{Limitations and future work.}
The benchmark covers one sign language, one humanoid, and only manual cues. Recognition BLEU and simulated execution do not establish Deaf comprehension. Execution is tested on only 20 motions in simulation with qualitative hardware replay. Future work will add non-manual cues, other robots and sign languages, dialogue-level interaction, and comprehension studies with Deaf signers.

\bibliography{iclr2027_conference}
\bibliographystyle{iclr2027_conference}
\clearpage
\appendix
\printappendixcontents
\enableappendixcontents

\FloatBarrier
\section{Data artifacts and provenance}
\label{app:data}

The appendix follows the order of the pipeline. Appendix~\ref{app:data} describes the stored motion versions, their time alignment, the repair records, and the evaluation subsets. Appendix~\ref{app:implementation} gives the construction and control settings, and Appendix~\ref{app:evaluation} defines the metrics and their aggregation. Appendix~\ref{app:additional} reports the complete control results, the threshold sensitivity, and the provenance-stratified analysis. Appendix~\ref{app:qualitative} collects geometry-repair cases, hardware replays, and human-repair cases.

\FloatBarrier
\subsection{Stored versions and time alignment}
\label{app:representations}
\label{app:alignment}
\label{app:shared_eval}

Each sequence is identified by its CSL-Daily video and sentence annotation and by the corresponding SOKE SMPL-X fit. Table~\ref{tab:representation_schema} lists what is stored for each version. Human versions store SMPL-X arrays at the native 30\,FPS. Robot versions store the floating base and 53 joint coordinates: 29 body coordinates and 24 hand coordinates, of which six per hand are actuated and the rest follow the finger coupling. Every robot version records its robot model, joint ordering, collision model, and coupling definitions. The Global smoothing output $\mathcal{H}^{\mathrm{global}}$ is stored as a processing baseline. Each version also records the configuration that produced it and whether generation succeeded.

\begin{table}[!htbp]
\caption{Stage-wise motion representations and linked records.}
\label{tab:representation_schema}
\centering
\footnotesize
\renewcommand{\tabularxcolumn}[1]{m{#1}}
\begin{tabularx}{\linewidth}{@{}l>{\raggedright\arraybackslash}X>{\raggedright\arraybackslash}X@{}}
\toprule
\multicolumn{1}{@{}c}{State} & \multicolumn{1}{c}{Motion payload} & \multicolumn{1}{c@{}}{Linked records} \\
\midrule
$\mathcal{H}^{\mathrm{raw}}$ & SMPL-X root, translation, shape, body, and hand arrays & Video and sentence IDs, source frames and timestamps, coordinate convention \\
$\mathcal{H}^{\mathrm{fused}}$ & Same human-motion schema & Part masks, repair intervals, anchors, operations, observation-support scores \\
$\mathcal{G}^{\mathrm{direct}}$ & Floating base and 53-coordinate robot motion & Robot model, joint ordering, collision model, coupling definitions \\
$\mathcal{G}^{\mathrm{repair}}$ & Same robot-motion schema & Deformation, residual penetration, constraint checks, certification flag \\
\bottomrule
\end{tabularx}
\end{table}

All versions share the source time axis. Missing observations, cropped intervals, and synthetic transitions are marked by masks rather than removed, so paired methods keep the same timestamps. Human metrics and geometry repair use the native 30\,Hz samples. Geometry evaluation, policy training, and replay use a common 50\,Hz grid with $t=j/50$ strictly below the final source timestamp $(T-1)/30$. On this grid, scalar joint coordinates are interpolated linearly and the floating-base orientation by SLERP; passive finger coordinates are then reconstructed from the coupling, and forward kinematics is recomputed. The human-to-robot part mapping carries the repair records of the left and right arm and hand to the corresponding robot joints. Synthetic transitions between clips carry a separate validity flag instead of a repair record.

\FloatBarrier
\subsection{Repair provenance}
\label{app:provenance}

For every frame and part group, human repair stores the operation that produced the pose and the fixed observation-support score in Table~\ref{tab:provenance}.

\begin{table}[!htbp]
\caption{Operation labels and fixed scores for human-motion repair provenance.}
\label{tab:provenance}
\centering
\begin{tabular}{@{}lr@{}}
\toprule
\multicolumn{1}{@{}c}{Provenance category} & \multicolumn{1}{c@{}}{Score} \\
\midrule
Retained source fit & 1.00 \\
Exit transition from a frozen segment & 0.70 \\
Two-anchor interpolation & 0.50 \\
Single-anchor freezing & 0.25 \\
Long-segment freezing & 0.10 \\
\bottomrule
\end{tabular}
\end{table}

The score states how directly an operation is supported by the video. It is neither an error probability nor a measure of sign quality: natural stillness and weak visual evidence can both lead to holding, so the frequency of an operation is not an error rate. Geometry deformation, residual penetration, and controller tracking are stored as separate fields. The high-support subset used in Appendix~\ref{app:provenance_stratified} keeps frames with a score above 0.5, that is, retained fits and exit transitions.

\FloatBarrier
\subsection{Construction coverage and evaluation subsets}
\label{app:production}
\label{app:distributions}
\label{app:costs}

Every source sequence stays in the manifest with a status for each stage: human-fit validity, human repair, direct retargeting, geometry repair, production validation, conversion to the 50\,Hz grid, and control readiness. A sequence that fails at one stage keeps its failure reason, so output, validation, and control coverage have separate denominators.

Of the 20,652 source sequences, 20,649 have repaired human motion (2,461,106 frames at 30\,FPS) and direct robot references for the Raw, Global, and Fused inputs. The three excluded sequences, given by their CSL-Daily identifiers, are \texttt{S000005\_P0004\_T00}, \texttt{S000007\_P0004\_T00}, and \texttt{S001818\_P0000\_T00}. Geometry repair yields 20,648 references that pass the production checks (2,460,979 frames). The remaining sequence, \texttt{S005379\_P0009\_T00}, fails the residual check of the quadratic program and has no numerical result; it counts as a failure in coverage and is excluded from paired deformation statistics. Human and Direct coverage is 99.985\%, and repaired coverage is 99.981\%. On the 50\,Hz grid, the geometry evaluation contains 4,074,112 frames and 12,160,392 interior samples per condition. Table~\ref{tab:subsets} lists the evaluation sets, which were selected before the experiment results were inspected.

\begin{table}[!htbp]
\caption{Data partitions used by the construction and evaluation pipeline.}
\label{tab:subsets}
\centering
\small
\renewcommand{\tabularxcolumn}[1]{m{#1}}
\begin{tabularx}{\linewidth}{@{}lrl>{\raggedright\arraybackslash}X@{}}
\toprule
\multicolumn{1}{@{}c}{Set} & \multicolumn{1}{c}{Size} & \multicolumn{1}{c}{Source} & \multicolumn{1}{c@{}}{Role} \\
\midrule
$\mathcal{D}_{\mathrm{all}}$ & 20,652 & Full source manifest & Production coverage, corpus statistics, and paired human and geometry evaluation \\
$\mathcal{D}_{\mathrm{diag}}$ & 600 & Fixed stratified subset & Repair-cue diagnostics (Figure~\ref{fig:crosstab}) \\
$\mathcal{D}_{\mathrm{content}}$ & 75 & Recognizer-held-out dev/test sequences & Content preservation (frozen recognizer) \\
$\mathcal{D}_{\mathrm{control}}$ & 20 & Fixed random sample & Self-target and fixed-reference replay \\
\bottomrule
\end{tabularx}
\end{table}

The 20 motions of $\mathcal{D}_{\mathrm{control}}$ were sampled at random from CSL-Daily and are fixed across all control conditions and seeds. Table~\ref{tab:control_semantics} lists their identifiers with English translations of the official sentence annotations.

\begingroup
\small
\setlength{\tabcolsep}{4pt}
\renewcommand{\arraystretch}{1.18}
\setlength{\LTleft}{0pt plus 1fill}
\setlength{\LTright}{0pt plus 1fill}
\setlength{\LTcapwidth}{\linewidth}
\begin{longtable}{@{}l>{\raggedright\arraybackslash}m{0.70\linewidth}@{}}
\caption{The 20 control motions with English translations of their CSL-Daily sentence annotations.}
\label{tab:control_semantics}\\
\toprule
\multicolumn{1}{@{}c}{Sequence ID} & \multicolumn{1}{c@{}}{Sentence-level meaning (English translation)} \\
\midrule
\endfirsthead
\multicolumn{2}{@{}l}{\small\itshape Table~\thetable\ continued}\\
\toprule
\multicolumn{1}{@{}c}{Sequence ID} & \multicolumn{1}{c@{}}{Sentence-level meaning (English translation)} \\
\midrule
\endhead
\midrule
\multicolumn{2}{r@{}}{\small Continued on next page}\\
\endfoot
\bottomrule
\endlastfoot
\texttt{S000374\_P0004\_T00} & I go to bed early too, at ten every night. Going to bed early is good for your health. \\
\texttt{S000443\_P0000\_T00} & Take a break. \\
\texttt{S000748\_P0004\_T00} & Take the bus to the train station from across the road. \\
\texttt{S000919\_P0008\_T00} & What tasty dishes does your restaurant have? Please recommend some. \\
\texttt{S001518\_P0004\_T00} & Why do you drink coffee at night? \\
\texttt{S001836\_P0000\_T00} & Only by delighting the eye and mind can works of art move and educate people. \\
\texttt{S001883\_P0000\_T00} & The enemy always overestimates its own strength. \\
\texttt{S002075\_P0000\_T00} & Maintain broad contact with the public; do not isolate yourself. \\
\texttt{S002219\_P0000\_T00} & This week, the company will announce my proposed paid-leave plan for low-income mothers and fathers. \\
\texttt{S002297\_P0000\_T00} & Strengthen composition teaching to improve students' written and oral expression. \\
\texttt{S003335\_P0000\_T00} & How could you forget so quickly what the teacher just explained? \\
\texttt{S003637\_P0001\_T00} & I saw many colorful flowers in the park, which were worthy of admiration. \\
\texttt{S003742\_P0001\_T00} & He has plenty of time to deal with this matter. \\
\texttt{S003956\_P0002\_T00} & The teacher severely criticized me for making a mistake. \\
\texttt{S004575\_P0008\_T00} & Feel proud of being able to help others. \\
\texttt{S004729\_P0003\_T00} & My phone was stolen on the subway. \\
\texttt{S004989\_P0007\_T00} & Do your own work well and ignore other people's ridicule. \\
\texttt{S005258\_P0006\_T00} & He was the only patient in the emergency room, so many doctors came. \\
\texttt{S005432\_P0009\_T00} & Sichuan opera, as its name suggests, is a local opera popular in Sichuan. \\
\texttt{S006044\_P0000\_T00} & Your PowerPoint presentation has no tables at all; it is all text. \\
\end{longtable}
\endgroup

\FloatBarrier
\section{Construction and control implementation}
\label{app:implementation}

\FloatBarrier
\subsection{Human-motion repair}
\label{app:human}
\label{app:repair_config}
\label{app:evidence}
\label{app:sensitivity}

The image cue uses the RTMPose keypoints of Section~\ref{sec:human_repair}: hand-keypoint confidence, wrist, elbow, and shoulder visibility, hand bounding boxes, blur, and temporal changes in keypoints and boxes. The motion cue uses changes and oscillations of the fitted rotations, with thresholds of 0.30\,rad for arms and 0.70\,rad for hands between adjacent 30\,FPS samples. Masks are formed separately for the left arm, left hand, right arm, and right hand and expanded by one frame before repair.

Anchors are selected from unflagged frames of the same part group. A gap of at most 10 frames with an anchor on each side is filled by spherical interpolation. If the resulting per-frame step would exceed 90\% of the group threshold, the right anchor moves one unflagged frame further at a time, up to twice the gap length, and the extension is recorded. A gap of at most 10 frames with a single anchor holds the available pose. A longer interior gap holds the preceding reliable pose and blends into the next reliable observation over its last four frames. A long gap at the start of a sequence holds the medoid pose of the flagged segment and blends into the first reliable frame in the same way, and a long gap at the end holds the last reliable pose. Gaps without any anchor are left unrepaired and marked as such. After repair, arm windows that still contain large steps receive light smoothing. A window is detected where the largest collar, shoulder, or elbow step exceeds the larger of 0.08\,rad and the sequence median plus three scaled median absolute deviations; it is expanded by five frames, smoothed in rotation space with a Gaussian of $\sigma=4$ frames, and blended into the unsmoothed motion with $\sigma=2$ frames. Wrist and finger rotations are not smoothed after repair.

The Global baseline applies Gaussian smoothing in rotation space to the whole sequence, with $\sigma=1.75$ frames for the body and arms and $\sigma=0.8$ frames for the hands. Root, translation, and timing are unchanged, and local repair is disabled, so Global represents sequence-wide smoothing.

\begin{figure}[!htbp]
\centering
\includegraphics[width=0.96\linewidth]{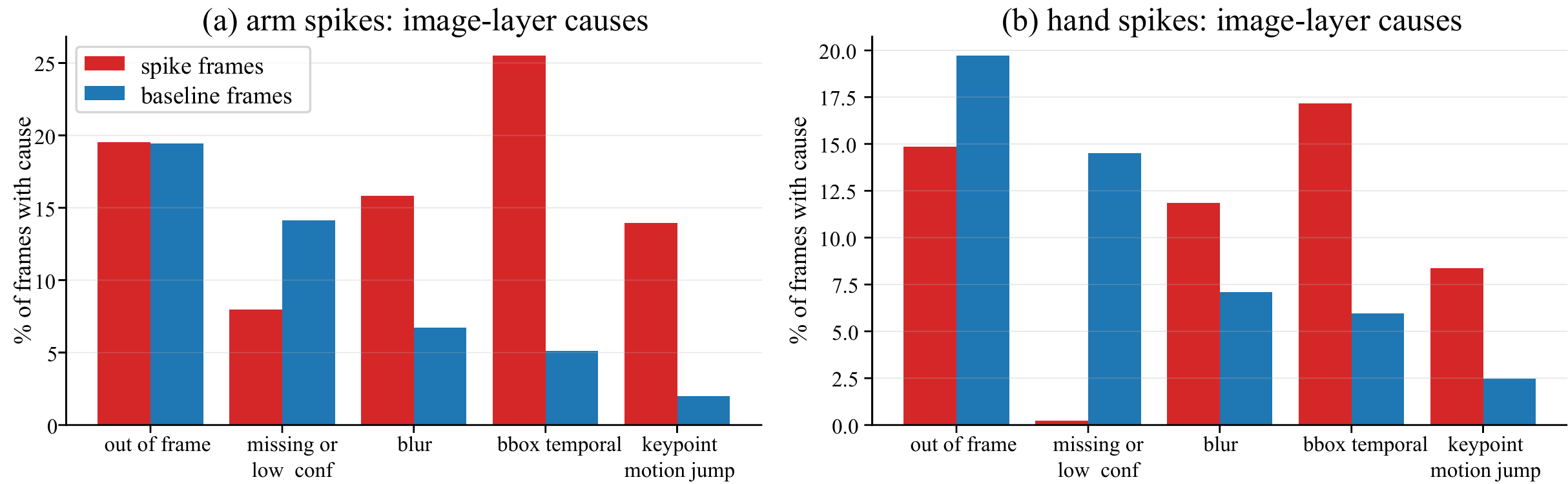}
\caption{Image-cue rates on frames flagged by the diagnostic motion rule (red) and on unflagged frames (blue) in the fixed 600-sequence diagnostic set.}
\label{fig:crosstab}
\end{figure}

Figure~\ref{fig:crosstab} shows that the two cues are complementary. On the fixed 600-sequence diagnostic set, it compares image-cue rates on frames flagged by a diagnostic motion rule, which uses the larger of the fixed threshold and the sequence median plus three scaled median absolute deviations, with the rates on unflagged frames. Out-of-frame observations are no more frequent on flagged arm frames (19.5\% in both) and less frequent on flagged hand frames (14.9\% versus 19.7\%); missing or low-confidence keypoints behave similarly. Blur, bounding-box instability, and keypoint jumps are more frequent on flagged frames. No image cue fires on about 53\% of flagged arm frames and 67\% of flagged hand frames, so neither cue covers the frames found by the other. The comparison shows co-occurrence, not cause.

\FloatBarrier
\subsection{Robot platform and collision model}
\label{app:platform}

\begin{figure}[!htbp]
\centering
\includegraphics[width=\linewidth]{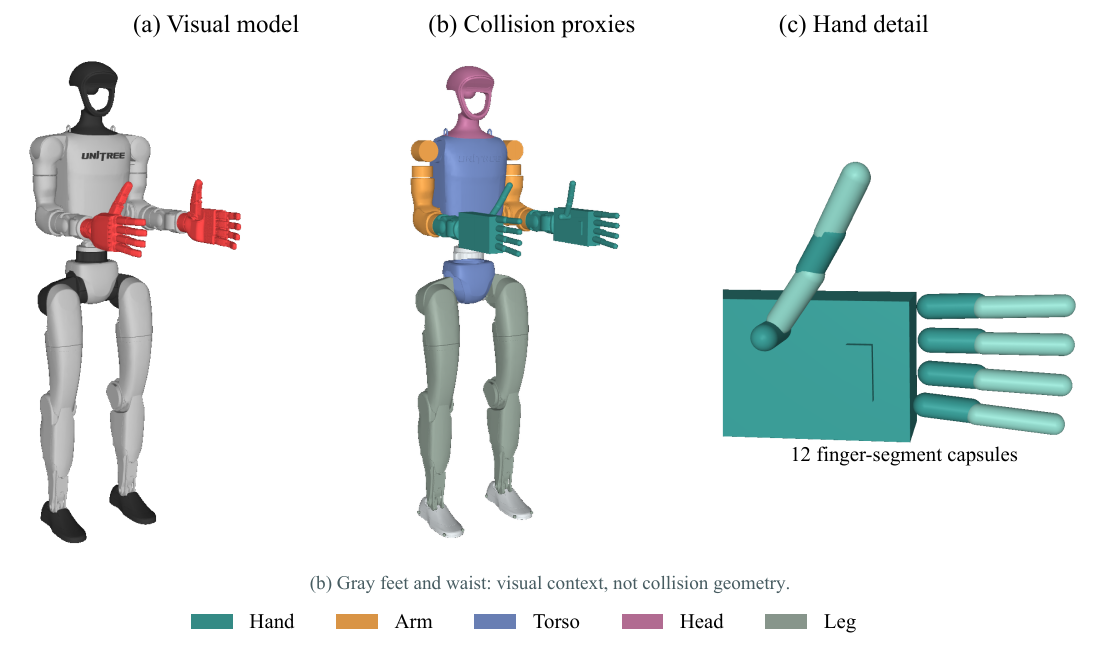}
\caption{Unitree G1 with Inspire RH56DFTP hands. (a) Visual model. (b) Collision proxies colored by group. (c) Hand detail with the twelve finger-segment capsules; alternating shades distinguish adjacent segments. All views are rendered from the production MuJoCo model.}
\label{fig:platform}
\end{figure}

All robot references target a Unitree G1 humanoid with two Inspire RH56DFTP hands (Figure~\ref{fig:platform}). The G1 body has 29 revolute degrees of freedom: 12 in the legs, 3 in the waist, and 7 in each arm (shoulder pitch, roll, and yaw; elbow; wrist roll, pitch, and yaw). Each hand has 6 actuated joints and 6 passive joints driven by linear couplings, so the assembled model has 53 revolute joints, 41 actuated coordinates, and 12 coupling constraints (Table~\ref{tab:platform}). The same model is used for retargeting, geometry repair, the collision audit, and simulated control.

\begin{table}[!htbp]
\caption{Kinematic structure of the G1 + RH56DFTP platform. Hand ranges are given per hand.}
\label{tab:platform}
\centering
\small
\setlength{\tabcolsep}{4pt}
\begin{tabularx}{\linewidth}{@{}lrX@{}}
\toprule
Group & DoF & Joints and ranges \\
\midrule
Legs & 12 & Hip pitch, roll, yaw; knee; ankle pitch, roll (per leg) \\
Waist & 3 & Yaw, roll, pitch \\
Arms & 14 & Shoulder pitch, roll, yaw; elbow; wrist roll, pitch, yaw (per arm) \\
Hands, actuated & 12 & Thumb rotation $[0,1.164]$\,rad; thumb flexion $[0,0.586]$\,rad; index, middle, ring, and little flexion $[0,1.438]$\,rad \\
Hands, passive & 12 & Distal thumb and finger joints coupled to the actuated joints \\
\midrule
Total & 53 & 41 actuated coordinates, 12 mimic constraints \\
\bottomrule
\end{tabularx}
\end{table}

\paragraph{Collision proxies.}
Signed distances between collision proxies are computed with the MuJoCo distance query. Proxies are grouped into left and right hand, arm, and leg, torso, and head. Body links use their collision meshes, with cylinders at the shoulders and four 5\,mm contact spheres per foot. Each hand has 17 finer proxies: three wrist-housing meshes, a palm box, a palm-sensor box, and twelve capsules (radius 7--8\,mm, half-length 12.5--25\,mm) that bound the finger segments. The coarse proxies capture contact between the hands and the bulky arm and torso casings, and the capsules capture the narrow gaps between fingers.

\paragraph{Audited pair set.}
$\mathcal{K}_{\mathrm{body}}$ pairs every hand proxy with every proxy of the other hand and of the arms, legs, torso, and head; $\mathcal{K}_{\mathrm{intra}}$ pairs the proxies within each hand. Three rules remove pairs whose contact is structural rather than a motion error: pairs on the same rigid assembly, pairs on adjacent articulated assemblies, and the two overlapping wrist-joint housings. The audit contains 1,519 pairs (Table~\ref{tab:collision_pairs}).

\begin{table}[!htbp]
\caption{Audited collision pairs of the G1 + RH56DFTP model.}
\label{tab:collision_pairs}
\centering
\small
\begin{tabular*}{\linewidth}{l@{\extracolsep{\fill}}lr}
\toprule
Set & Category & Pairs \\
\midrule
\multirow{6}{*}{$\mathcal{K}_{\mathrm{body}}$} & Inter-hand & 289 \\
& Hand-contralateral arm & 136 \\
& Hand-same-side arm & 134 \\
& Hand-torso & 102 \\
& Hand-head & 34 \\
& Hand-leg & 612 \\
$\mathcal{K}_{\mathrm{intra}}$ & Within each hand & 212 \\
\midrule
$\mathcal{K}_{\mathrm{eval}}$ & Total audited & 1,519 \\
Excluded & Adjacent assemblies (54), same rigid assembly (6), wrist housings (2) & 62 \\
\bottomrule
\end{tabular*} 
\end{table}

\FloatBarrier
\subsection{Robot retargeting and geometry repair}
\label{app:solver}

General Motion Retargeting supplies the body and global wrist motion, and dex-retargeting supplies the wrist-local handshape. Their outputs are combined with common joint clamping and the finger coupling to obtain $\mathcal{G}^{\mathrm{direct}}$. Geometry repair takes this trajectory as input. The arm stage fixes the root, legs, waist, and fingers and optimizes the 14 shoulder, elbow, and wrist coordinates to remove inter-hand and hand-body penetration. The finger stage then fixes everything except the 12 active finger joints and removes intra-hand penetration within a per-joint budget.

\paragraph{Arm stage.}
Let $\boldsymbol{x}_t\in\mathbb{R}^{14}$ be the arm coordinates, $\boldsymbol{x}^{(0)}$ their projected initialization, and $\mathcal{E}_{\mathrm{track}}$ and $\mathcal{E}_{\mathrm{rel}}$ the terms of Equations~\ref{eq:arm_terms} and~\ref{eq:bilateral_relation_main}, with $W_b=\operatorname{diag}(10I_3,5I_3)$ for shoulders and elbows and $\operatorname{diag}(10I_3,20I_3)$ for wrists (position, then rotation). Let $\boldsymbol{r}_t$ stack the two wrist pose residuals and $W_c=\operatorname{diag}(100I_3,20I_3,100I_3,20I_3)$. The wrist-residual continuity term is
\begin{equation}
\label{eq:wrist_residual_continuity}
\mathcal{E}_{\mathrm{cont}}(\boldsymbol{x})=\sum_{t=0}^{N-2}\lVert W_c(\boldsymbol{r}_{t+1}-\boldsymbol{r}_t)\rVert_2^2+\sum_{t=0}^{N-3}\lVert2W_c(\boldsymbol{r}_{t+2}-2\boldsymbol{r}_{t+1}+\boldsymbol{r}_t)\rVert_2^2,
\end{equation}
and the full arm-stage problem summarized by Equation~\ref{eq:geometry_repair} is
\begin{equation}
\label{eq:geometry_repair_full}
\begin{gathered}
\min_{\boldsymbol{x},\boldsymbol{s}}\quad
\sum_t\!\left(\mathcal{E}_{\mathrm{track}}(\boldsymbol{x}_t)
+\lambda_r\mathcal{E}_{\mathrm{rel}}(\boldsymbol{x}_t)\right)
+\lambda_v\lVert D_1\boldsymbol{x}\rVert_2^2
+\lambda_a\lVert D_2\boldsymbol{x}\rVert_2^2
+\lambda_j\lVert D_3\boldsymbol{x}\rVert_2^2 \\
+\lambda_c\mathcal{E}_{\mathrm{cont}}(\boldsymbol{x})
+\lambda_0\lVert\boldsymbol{x}-\boldsymbol{x}^{(0)}\rVert_2^2
+\lambda_s\lVert\boldsymbol{s}\rVert_2^2 \\
\text{s.t.}\quad
\boldsymbol{x}_{\min}\leq\boldsymbol{x}_t\leq\boldsymbol{x}_{\max},\quad
|D_1\boldsymbol{x}|\leq\boldsymbol{v}\Delta t,\quad
|D_2\boldsymbol{x}|\leq\boldsymbol{a}\Delta t^2,\quad
|D_3\boldsymbol{x}|\leq\boldsymbol{j}\Delta t^3, \\
\widehat d^{(\ell)}_{k,\tau}(\boldsymbol{x})+s_{k,\tau}\geq\gamma,
\quad s_{k,\tau}\geq0,\quad
(k,\tau)\in\mathcal{A}^{(\ell)}\subseteq\mathcal{K}_{\mathrm{body}}\times\Omega_M.
\end{gathered}
\end{equation}
The trajectory is optimized at 30\,Hz with $\Delta t=1/30$\,s. For $\tau=i+m/M\in\Omega_M$ with $i\in\{0,\ldots,N-2\}$ and $m\in\{0,\ldots,M\}$, $\boldsymbol{x}(\tau)=(1-m/M)\boldsymbol{x}_i+(m/M)\boldsymbol{x}_{i+1}$. At iteration $\ell$, the active set $\mathcal{A}^{(\ell)}$ contains the pairs closer than 8\,mm, and $\widehat d^{(\ell)}_{k,\tau}=d_k(\boldsymbol{q}^{(\ell)}_\tau)+(\boldsymbol{g}^{(\ell)}_{k,\tau})^\top(\boldsymbol{x}(\tau)-\boldsymbol{x}^{(\ell)}(\tau))$ linearizes the signed distance with its gradient $\boldsymbol{g}^{(\ell)}_{k,\tau}$ with respect to the arm coordinates. The clearance is $\gamma=3$\,mm for every pair. Table~\ref{tab:geometry_weights} lists the coefficients; because the terms keep their physical units, the coefficients are scalings rather than dimensionless trade-offs. The speed, acceleration, and jerk limits are 4\,rad/s, 45\,rad/s$^2$, and 300\,rad/s$^3$ for shoulders and elbows and 5.4\,rad/s, 70\,rad/s$^2$, and 500\,rad/s$^3$ for wrists.

\begin{table}[!htbp]
\caption{Geometry-repair objective and metric scalings. Squared implementation weights are shown as their effective coefficients in Equation~\ref{eq:geometry_repair_full}.}
\label{tab:geometry_weights}
\centering
\small
\setlength{\tabcolsep}{4pt}
\begin{tabularx}{\linewidth}{@{}lrlX@{}}
\toprule
Quantity & Value & Implementation value & Role \\
\midrule
$\lambda_r$ & $144$ & $w_r=12$ & Inter-wrist displacement residual \\
$\lambda_v$ & $0.3$ & $0.3$ & First joint difference \\
$\lambda_a$ & $8$ & $8$ & Second joint difference \\
$\lambda_j$ & $100$ & $100$ & Third joint difference \\
$\lambda_c$ & $1$ & $1$ & Wrist-residual continuity (Equation~\ref{eq:wrist_residual_continuity}) \\
$\lambda_0$ & $0.02$ & $0.02$ & Deviation from the projected initialization \\
$\lambda_s$ & $2.25{\times}10^8$ & $w_s=15000$ & Collision slack and nonlinear hinge violation \\
$W_c$ position/rotation & $100/20$ & $100/20$ & Wrist-residual continuity; second difference uses $2W_c$ \\
Tracking position/rotation & $10/5$ & $10/5$ & Shoulder and elbow pose residuals \\
Wrist position/rotation & $10/20$ & $10/(5{\times}4)$ & Wrist pose residuals \\
\bottomrule
\end{tabularx}
\end{table}

Each iteration solves a quadratic program in which a $10^{-4}I$ diagonal term stabilizes each per-frame block, and a step is accepted only after the nonlinear objective and all pairs of $\mathcal{K}_{\mathrm{body}}$ are evaluated. An adaptive trust region and separated seeds handle locally trapped configurations. The first attempt starts from the projected direct reference; later attempts may start from the preceding solution or a separated feasible pose. The default solve uses $M=8$, a failed candidate is retried with $M=16$ or $32$, and the validator uses $\max(16,2M)$ subdivisions. The validator checks inter-hand and hand-body geometry, joint ranges, speed, acceleration, jerk, finger coupling, quaternion norms, and finite values. The relation term is part of the objective, not an acceptance test. Candidates that fail validation stay in the output record but do not enter $\mathcal{G}^{\mathrm{repair}}$.

\paragraph{Finger stage.}
The finger stage optimizes the 12 active finger joints $\boldsymbol{q}^f$ (thumb rotation, thumb flexion, and index, middle, ring, and little flexion on each hand) and holds all other coordinates fixed. The 24 finger coordinates follow the linear coupling $\Phi\in\mathbb{R}^{24\times12}$ as $\Phi\boldsymbol{q}^f$. With $\Delta\boldsymbol{q}^f=\boldsymbol{q}^f-\boldsymbol{q}^{f,(0)}$ the change from the direct handshape, the full form of Equation~\ref{eq:finger_repair} is
\begin{equation}
\label{eq:finger_micro_full}
\begin{gathered}
\min_{\Delta \boldsymbol{q}^f,\boldsymbol{s}}\quad
\sum_{t=0}^{N-1}\lVert\Delta\boldsymbol{q}^f_t\rVert_2^2
+\beta_1\lVert D_1\Delta\boldsymbol{q}^f\rVert_2^2
+\beta_2\lVert D_2\Delta\boldsymbol{q}^f\rVert_2^2
+\beta_3\lVert D_3\Delta\boldsymbol{q}^f\rVert_2^2
+\rho^2\sum_{(k,\tau)\in\mathcal{A}^{(\ell)}}s_{k,\tau}^2 \\
\text{s.t.}\quad
|\Delta q^f_{t,j}|\leq\theta_{\max},\quad
\boldsymbol{q}^f_{\min}\leq\boldsymbol{q}^f_t\leq\boldsymbol{q}^f_{\max},\quad
|D_d\boldsymbol{q}^f|\leq\boldsymbol{u}_d, \\
\widehat d^{(\ell)}_{k,\tau}(\boldsymbol{q}^f)+s_{k,\tau}\geq\gamma_f,\quad
s_{k,\tau}\geq0,\quad
(k,\tau)\in\mathcal{A}^{(\ell)}\subseteq\mathcal{K}_f\times\Omega_M.
\end{gathered}
\end{equation}
$\mathcal{K}_f$ contains every audited pair whose distance depends on $\boldsymbol{q}^f$, namely intra-hand pairs and inter-hand or hand-body pairs with a finger segment. A pair enters $\mathcal{A}^{(\ell)}$ when its distance is below 3\,mm, and its linearization is $\widehat d^{(\ell)}_{k,\tau}(\boldsymbol{q}^f)=d_k(\boldsymbol{q}^{f,(\ell)}_\tau)+(\boldsymbol{g}^{(\ell)}_{k,\tau})^\top\Phi(\boldsymbol{q}^f(\tau)-\boldsymbol{q}^{f,(\ell)}(\tau))$, where $\boldsymbol{g}^{(\ell)}_{k,\tau}$ is the gradient of $d_k$ with respect to the 24 finger coordinates. The weights $\beta_1,\beta_2,\beta_3$ take the place of $\lambda_v,\lambda_a,\lambda_j$ in the $\mathcal{E}_{\mathrm{smooth}}$ term of Equation~\ref{eq:finger_repair}. The derivative bounds are $\boldsymbol{u}_d=\max(u_{\max}^d\Delta t^d,|D_d\boldsymbol{q}^{f,(0)}|)+10^{-10}$ for $d=1,2,3$, with $u_{\max}^1=v_{\max}^f$, $u_{\max}^2=a_{\max}^f$, and $u_{\max}^3=j_{\max}^f$ from Table~\ref{tab:finger_repair_params}, so derivatives already present in the direct reference are not amplified. Once all slack is zero, the solver continues to minimize $\sum_t\lVert\Delta\boldsymbol{q}^f_t\rVert_2^2$, which returns the fingers toward the direct handshape. Table~\ref{tab:finger_repair_params} lists the parameters.

\begin{table}[!htbp]
\caption{Implementation parameters for the finger stage (Equation~\ref{eq:finger_micro_full}).}
\label{tab:finger_repair_params}
\centering
\small
\setlength{\tabcolsep}{4pt}
\begin{tabularx}{\linewidth}{@{}lrlX@{}}
\toprule
Parameter & Value & Unit / Dimension & Role \\
\midrule
$\theta_{\max}$ & $20.0^\circ$ ($0.34907$) & rad & Hard per-joint budget on active finger joints \\
$\gamma_f$ & $0.2$ & mm ($2\times 10^{-4}$\,m) & Target geometric clearance margin \\
$\beta_1$ & $0.5$ & dimensionless & First difference regularizer ($D_1$) \\
$\beta_2$ & $5.0$ & dimensionless & Second difference regularizer ($D_2$) \\
$\beta_3$ & $25.0$ & dimensionless & Third difference regularizer ($D_3$) \\
$\rho$ & $10{,}000$ & $\rho^2 = 10^8$ & Quadratic penalty on penetration slack $s_{k,\tau}$ \\
$v_{\max}^f$ & $4.8$ & rad/s & Active finger angular velocity upper bound \\
$a_{\max}^f$ & $100.0$ & rad/s$^2$ & Active finger angular acceleration upper bound \\
$j_{\max}^f$ & $2000.0$ & rad/s$^3$ & Active finger angular jerk upper bound \\
Trust region step & $0.15$ & rad & Maximum coordinate step per SQP iteration \\
Subdivision schedule & $\{4, 8, 16\}$ & intervals/frame & Progressive continuous collision temporal refinement \\
Iteration limits & $(24, 32, 48)$ & iterations & Max SQP iterations for subdivision rounds $M\in\{4,8,16\}$ \\
\bottomrule
\end{tabularx}
\end{table}

A sequence is certified when it satisfies Equation~\ref{eq:certified} on both the 30\,Hz grid with 16 subdivisions and the 50\,Hz grid with three interior samples. The 50\,Hz check is needed because resampling at a non-integer ratio exposes poses between the 30\,Hz samples that the native check does not cover. All 20,648 sequences of $\mathcal{G}^{\mathrm{repair}}$ also pass the production checks, and the finger stage leaves the root, waist, and arm coordinates of the arm stage unchanged. Every uncertified sequence completed all three rounds of the subdivision schedule. For the 2,434 uncertified sequences, $\mathcal{G}^{\mathrm{repair}}$ stores the final within-budget result together with per-pair residual records (Appendix~\ref{app:intrahand_results}). The collision model, pair set, $1\,\mu$m tolerance, and $20^\circ$ budget are the same for all sequences.

\FloatBarrier
\subsection{Reference-tracking policy and curriculum}
\label{app:control}

Following BeyondMimic, each policy of our controller is trained on one reference motion. All our control conditions share the residual tracking policy, observations, physics, termination rules, and PPO settings. The actor and critic are ELU MLPs with hidden widths 512, 256, and 128. Training uses 1,024 parallel environments, 24 rollout steps per iteration, two epochs, four minibatches, a fixed learning rate of $3\times10^{-5}$, discount 0.99, GAE parameter 0.95, clip ratio 0.2, entropy coefficient $2.5\times10^{-4}$, value coefficient 0.5, maximum gradient norm 0.5, and initial action noise 0.20. Actor and critic inputs are normalized empirically. The simulator uses a 0.005\,s step with four steps per 50\,Hz control step.

The arm joints, and in the second stage the active finger joints, receive residual actions around their reference targets, clipped to $\pm0.10$\,rad. Each action is scaled per joint by a quarter of the effort limit divided by the stiffness. Finger actions are disabled in the first stage and scaled by 0.02 in the second. Body-joint PD gains follow BeyondMimic, with stiffness $I_a\omega^2$ and damping $2\zeta I_a\omega$ for reflected armature $I_a$, $\omega=2\pi\cdot10$\,rad/s, and $\zeta=2$; the hand joints use a stiffness of 5\,N\,m/rad and a damping of 0.2\,N\,m\,s/rad. Each episode starts at a time drawn uniformly over the reference, with small perturbations of the root pose (at most 2\,cm horizontally, 5\,mm vertically, and 0.05\,rad in orientation) and of the joint positions (at most 0.03\,rad). Episodes last at most 10\,s in the first stage and 14\,s in the second and end early under the termination rules of the replay (Appendix~\ref{app:control_metrics}). Domain randomization, external pushes, and observation noise are disabled.

The reward per control step is
\begin{equation}
\label{eq:control_reward}
R_t=\Delta t_c\sum_{k\in\mathcal{R}}w_k r_k(t),
\qquad \Delta t_c=0.02\ \mathrm{s}.
\end{equation}
For a set $\mathcal{A}$ of tracked bodies, vector and rotation errors and the tracking kernel are
\begin{equation}
\label{eq:control_reward_kernel}
\begin{gathered}
E_{\mathrm{vec}}
=\frac{1}{|\mathcal{A}|}\sum_{a\in\mathcal{A}}\lVert
\boldsymbol{y}_{a,t}-\boldsymbol{y}^{\mathrm{ref}}_{a,t}\rVert_2^2,\qquad
E_{\mathrm{rot}}
=\frac{1}{|\mathcal{A}|}\sum_{a\in\mathcal{A}}
d_{SO(3)}(R_{a,t},R^{\mathrm{ref}}_{a,t})^2,\\
r_{\mathrm{trk}}(E;\sigma)=\exp(-E/\sigma^2).
\end{gathered}
\end{equation}
Joint errors $E_q$ use the same mean-squared form over joints, and anchor terms have $|\mathcal{A}|=1$. Table~\ref{tab:control_rewards} lists every nonzero term of the two training stages. Single-stage runs use the Stage-2 column throughout; two-stage runs use Stage 1 for the first 1,500 iterations and Stage 2 for the remaining 1,500. The indicator $z_t$ marks frames of a preparation segment. The benchmark references contain only the signing motion, so $z_t=0$ and the four ready terms are inactive in all reported runs. No torque or joint-acceleration penalty is used.

\begin{table}[!htbp]
\caption{Reward terms used by the reported Stage-1 and Stage-2 tasks. Tracking rows use $r_{\mathrm{trk}}$ from Equation~\ref{eq:control_reward_kernel}. A dash denotes an inactive term. The contact term counts configured bodies whose peak force in the sensor history exceeds 1\,N.}
\label{tab:control_rewards}
\centering
\footnotesize
\setlength{\tabcolsep}{3pt}
\begin{tabularx}{\linewidth}{@{}lXrrr@{}}
\toprule
Term & Unweighted term $r_k$ & $\sigma$ & $w_k^{(1)}$ & $w_k^{(2)}$ \\
\midrule
Anchor position & $r_{\mathrm{trk}}(\lVert\boldsymbol{p}-\boldsymbol{p}^{\mathrm{ref}}\rVert_2^2;\sigma)$ & 0.50\,m & 1.00 & 1.00 \\
Anchor orientation & $r_{\mathrm{trk}}(d_{SO(3)}^2;\sigma)$ & 0.60\,rad & 0.75 & 0.75 \\
Body position & $r_{\mathrm{trk}}(E_{\mathrm{vec}};\sigma)$, 14 bodies & 0.35\,m & 3.00 & 3.00 \\
Body orientation & $r_{\mathrm{trk}}(E_{\mathrm{rot}};\sigma)$, 14 bodies & 0.60\,rad & 1.00 & 1.00 \\
Body linear velocity & $r_{\mathrm{trk}}(E_{\mathrm{vec}};\sigma)$, 14 bodies & 1.00\,m/s & 0.25 & 0.25 \\
Body angular velocity & $r_{\mathrm{trk}}(E_{\mathrm{vec}};\sigma)$, 14 bodies & 3.14\,rad/s & 0.25 & 0.25 \\
Upper-body joint position & $r_{\mathrm{trk}}(E_q;\sigma)$, 17 joints & 0.40\,rad & 1.75 & 1.75 \\
Leg pose prior & $r_{\mathrm{trk}}(E_q;\sigma)$, 12 joints & 0.60\,rad & 0.25 & 0.25 \\
Default hand pose & $r_{\mathrm{trk}}(E_q;\sigma)$, 24 joints & 0.50\,rad & 0.25 & -- \\
Wrist orientation & $r_{\mathrm{trk}}(E_{\mathrm{rot}};\sigma)$, two wrists & 0.45\,rad & -- & 0.30 \\
Active-finger position & $r_{\mathrm{trk}}(E_q;\sigma)$, 12 joints & 0.35\,rad & -- & 0.40 \\
Active-finger velocity & $r_{\mathrm{trk}}(E_{\dot q};\sigma)$, 12 joints & 6.00\,rad/s & -- & 0.05 \\
Action rate & $\lVert\boldsymbol{a}_t-\boldsymbol{a}_{t-1}\rVert_2^2$ & -- & $-0.03$ & $-0.03$ \\
Joint limit & $\min\!\left(10,\sum_j[\ell_j-q_j]_+ + [q_j-u_j]_+\right)$ & -- & $-0.25$ & $-0.25$ \\
Undesired contact & Number of force-threshold violations & -- & $-0.10$ & $-0.10$ \\
Flat orientation & $\lVert\boldsymbol{g}^{b}_{xy}\rVert_2^2$ & -- & $-2.00$ & $-2.00$ \\
Base angular velocity & $\lVert\boldsymbol{\omega}^{b}_{xy}\rVert_2^2$ & -- & $-0.50$ & $-0.50$ \\
Base linear velocity & $\lVert\boldsymbol{v}^{b}_{xy}\rVert_2^2$ & -- & -- & $-0.20$ \\
Ready pose & $z_t r_{\mathrm{trk}}(E_{q,29};0.35)$ & 0.35\,rad & -- & 1.00 \\
Ready joint velocity & $z_t\sum_{j\in\mathcal{J}_{29}}\dot q_j^2$ & -- & -- & $-0.10$ \\
Ready base linear velocity & $z_t\lVert\boldsymbol{v}^{b}\rVert_2^2$ & -- & -- & $-0.20$ \\
Ready base angular velocity & $z_t\lVert\boldsymbol{\omega}^{b}\rVert_2^2$ & -- & -- & $-0.20$ \\
\bottomrule
\end{tabularx}
\end{table}

The 14 tracked bodies are the pelvis, torso, and the left and right hip-roll, knee, ankle-roll, shoulder-roll, elbow, and wrist-yaw links. The upper-body joint set contains the three waist joints and the 14 arm joints, and the ready set adds the 12 active finger joints. The leg prior covers the 12 leg joints. The undesired-contact term excludes the ankle-roll, wrist, hand, base, and force-sensor links.

The actor receives $\boldsymbol{o}^{\pi}_t\in\mathbb{R}^{274}$ and the critic $\boldsymbol{o}^{V}_t\in\mathbb{R}^{406}$ (Table~\ref{tab:control_observations}). All blocks come from the current control step, with no history, lookahead, phase clock, or observation noise. Orientations use the first two columns of the rotation matrix. The 12 passive finger coordinates keep their slots in the measured joint blocks but are set to zero, because their values follow from the active joints.

\begin{table}[!htbp]
\caption{Actor and critic observation spaces. A dash denotes a block absent from that input. Reference position and velocity are the current 53-coordinate targets on the 50\,Hz control grid.}
\label{tab:control_observations}
\centering
\small
\setlength{\tabcolsep}{4pt}
\begin{tabularx}{\linewidth}{@{}Xrr@{}}
\toprule
Observation block & Actor & Critic \\
\midrule
Reference joint positions and velocities & 106 & 106 \\
Reference-anchor relative position & -- & 3 \\
Reference-anchor relative orientation & 6 & 6 \\
Tracked-body positions ($14\times3$) & -- & 42 \\
Tracked-body orientations ($14\times6$) & -- & 84 \\
Base linear velocity & -- & 3 \\
Base angular velocity & 3 & 3 \\
Joint position relative to default & 53 & 53 \\
Joint velocity relative to default & 53 & 53 \\
Previous action & 53 & 53 \\
\midrule
Total & \textbf{274} & \textbf{406} \\
\bottomrule
\end{tabularx}
\end{table}

Two-stage training tracks the body, arms, and wrists with the default handshape for 1,500 iterations, then restores the policy and observation normalizer and adds finger tracking for another 1,500. Every condition, including single-stage training, rebuilds its optimizer at iteration 1,500, so all conditions use 73,728,000 environment steps per policy. In the second stage, the action reference of active finger joint $j$ is ramped from the default pose $q_j^0$ to the reference:
\begin{equation}
\label{eq:finger_curriculum}
\begin{gathered}
\bar q_{n,j}=q_j^0+\alpha(n)\bigl(q^{\mathrm{ref}}_{t(n),j}-q_j^0\bigr),
\qquad
\alpha(n)=\operatorname{clip}\!\left(\frac{n-h}{r},0,1\right),\\
h=\max(1,\lfloor0.035T\rfloor),\qquad r=\max(10,\lfloor0.21T\rfloor).
\end{gathered}
\end{equation}
Here $n$ counts action updates since the episode reset, $t(n)$ is the corresponding reference frame, and $T$ is the number of reference frames on the control grid. All training references have $61\leq T\leq481$, so $h+r<T$. The residual action is added to $\bar q_{n,j}$ before joint-limit clipping and passive-finger reconstruction, while observations and finger rewards use the full reference throughout. Because training episodes start at random reference times, the ramp restarts at every episode reset; evaluation applies the full finger reference from the first frame. The no-ramp condition shares the first-stage policy and applies the full finger reference at once.

\paragraph{External AMP baseline.}
We adapt the public ProtoMotions 3.1 implementation of AMP to the same robot model and to the Raw + Direct, Fused + Direct, and Fused + Geometry references. For each reference track and seed, one policy is trained on all 20 motions with 64 parallel environments, 32 rollout steps, batch size 2,048, a discriminator replay buffer of 25,000, discriminator batch size 2,048, and a full-body discriminator. Training uses the IsaacGym backend of this implementation, with 36,000 updates or 73,728,000 environment steps per run. As for our controller, the trained policies are scored in the common IsaacLab replay of Appendix~\ref{app:control_metrics}. The AMP action path clips coupled finger targets to the hard joint limits, whereas the scorer reconstructs the soft-limit coupling; this changes one passive right-thumb reference in two of the 20 motions by at most 0.0205\,rad.

\FloatBarrier
\section{Metric definitions and evaluation protocols}
\label{app:evaluation}

\FloatBarrier
\subsection{Human-motion continuity and content preservation}
\label{app:human_metrics}
\label{app:content}

Human-motion metrics are computed per sequence at 30\,Hz and then averaged over the 20,649 paired sequences. On each side, the arm group contains the collar, shoulder, elbow, and wrist, and the hand group the fifteen finger joints. A step is abnormal when the largest within-group rotation exceeds 0.30\,rad for arms or 0.70\,rad for hands (Equation~\ref{eq:abnormal_rate}).

Table~\ref{tab:human_effects} gives the paired effects. The abnormal steps that remain after Fused repair lie mostly at the edges of the repair mask. For side $s$, the step from frame $t$ to $t+1$ is a mask edge when $m_t^s\ne m_{t+1}^s$, using the mask recorded before repair. Over the \BoundaryEligible{} sequences with at least one edge (\BoundaryEdges{} edges), 1.22\% of Fused edges exceed a threshold. All 7,730 of them are arm steps, split evenly between entry and exit (3,865 each). When a short segment is bridged by spherical interpolation, a large rotation between its anchors is spread over the interpolated steps, and the steps into and out of the segment can still exceed the arm threshold.

\begin{table}[!htbp]
\caption{Paired continuity effects over 20,649 sequences. Entries are mean differences with 95\% sequence-bootstrap intervals; pp denotes percentage points.}
\label{tab:human_effects}
\centering
\small
\setlength{\tabcolsep}{4pt}
\begin{tabular*}{\linewidth}{l@{\extracolsep{\fill}}cc}
\toprule
Metric & Fused$-$Raw & Fused$-$Global \\
\midrule
Arm abnormal steps (pp) & $-18.725\;[-18.823, -18.627]$ & $-1.143\;[-1.175, -1.111]$ \\
Hand abnormal steps (pp) & $-10.554\;[-10.619, -10.488]$ & $-0.140\;[-0.154, -0.126]$ \\
\bottomrule
\end{tabular*}
\end{table}

Tables~\ref{tab:human_diagnostics} and~\ref{tab:deviation_support} report how much each method changes the motion. Each unit is one side-frame of the arm or hand group, and its source deviation is the largest within-group rotation distance from Raw; a unit counts as changed when this distance exceeds 0.01\,rad (0.57$^\circ$). The stored repair mask covers 67.3\% of side-frames. Outside it, Fused leaves every hand unit unchanged and moves arm units by 0.042\,rad (2.4$^\circ$) on average, whereas Global moves nearly every unit, by 0.077\,rad for arms and 0.059\,rad for hands. The Fused arm changes outside the mask lie mostly within four frames of a flagged frame, where exit transitions blend back into the observed motion; 4.3\% of these units exceed 0.2\,rad. Inside the mask, Fused replaces the source pose with an interpolated or held pose, at mean deviations of 0.525\,rad for arms and 0.555\,rad for hands. Averaged over all units, Fused therefore changes fewer units than Global (79.75\% versus 99.68\% for arms and 61.34\% versus 85.17\% for hands) with larger means (0.3741 and 0.3807\,rad), and 96.4\% of its arm deviation and all of its hand deviation come from replaced flagged frames.

The path ratio divides the cumulative rotation-path length of each sequence by that of Raw, pooling joint steps over both sides; a pair with both lengths at most $10^{-9}$\,rad has ratio 100\%. For side $s$ with largest joint step $a_t^s$, the step variation is $30^3\operatorname{mean}_{t,s}|a_{t+2}^s-2a_{t+1}^s+a_t^s|$, which measures changes in step size rather than angular jerk. Fused shortens rotation paths to 36.49\% (arms) and 34.23\% (hands) of Raw and lowers hand step variation from 7,495.1 to 1,371.9, below Global's 2,722.4. Because a shorter path could also reflect removed articulation, content preservation is tested separately with the frozen recognizer.

\begin{table}[!htbp]
\caption{Supplementary motion diagnostics on all 20,649 pairs. Path ratio is rotation-path length relative to Raw; source deviation is measured from Raw.}
\label{tab:human_diagnostics}
\centering
\small
\setlength{\tabcolsep}{3pt}
\begin{tabular*}{\linewidth}{l@{\extracolsep{\fill}}rrrr}
\toprule
Group / condition & \makecell{Changed\\units (\%)} & \makecell{Path\\ratio (\%)} & \makecell{Step\\variation} & \makecell{Source dev.\\(rad)} \\
\midrule
Arms / Raw & 0.00 & 100.00 & 5618.9 & 0.0000 \\
Arms / Global smoothing & 99.68 & 47.21 & 585.4 & 0.1391 \\
Arms / Fused repair & 79.75 & 36.49 & 1037.3 & 0.3741 \\
\midrule
Hands / Raw & 0.00 & 100.00 & 7495.1 & 0.0000 \\
Hands / Global smoothing & 85.17 & 58.95 & 2722.4 & 0.1134 \\
Hands / Fused repair & 61.34 & 34.23 & 1371.9 & 0.3807 \\
\bottomrule
\end{tabular*}
\end{table}

\begin{table}[!htbp]
\caption{Source deviation from Raw split by the stored repair mask. Supported side-frames lie outside the mask (32.7\% of side-frames) and flagged side-frames inside it (67.3\%). Deviation means and changed shares pool side-frames over all 20,649 sequences; the last column is the share of the sequence-averaged source deviation in Table~\ref{tab:human_diagnostics} that comes from flagged side-frames.}
\label{tab:deviation_support}
\centering
\small
\setlength{\tabcolsep}{3pt}
\begin{tabular*}{\linewidth}{l@{\extracolsep{\fill}}rrrr}
\toprule
Group / condition & \makecell{Supported\\dev.\ (rad)} & \makecell{Supported\\changed (\%)} & \makecell{Flagged\\dev.\ (rad)} & \makecell{From\\flagged (\%)} \\
\midrule
Arms / Global smoothing & 0.0769 & 99.62 & 0.1632 & 82.5 \\
Arms / Fused repair & 0.0418 & 38.94 & 0.5246 & 96.4 \\
\midrule
Hands / Global smoothing & 0.0594 & 99.57 & 0.1364 & 83.5 \\
Hands / Fused repair & 0.0000 & 0.00 & 0.5549 & 100.0 \\
\bottomrule
\end{tabular*}
\end{table}

The content-preservation evaluation applies one frozen Uni-Sign model to Raw, Global, and Fused on the same 75 sequences, 38 official development and 37 official test examples excluded from recognizer training. The three conditions share the pose adapter, alignment fitted on Raw, FP32 inference, beam size four, and character-level text. BLEU uses the standard 13a tokenizer. Table~\ref{tab:bleu_orders} gives cumulative corpus BLEU with brevity penalty for all four orders. Paired intervals use 10,000 sequence resamples with seed 20260916 and recompute corpus BLEU from the resampled statistics. Table~\ref{tab:bleu_effects} reports pointwise 95\% intervals and 98.75\% intervals with a Bonferroni adjustment over the four orders within each contrast. Under this adjustment, Fused exceeds Global on BLEU-1 and BLEU-2, and Global stays below Raw on BLEU-1 and BLEU-2. Every Fused$-$Raw interval contains zero and lies within $\pm2.6$ BLEU points, so the recognizer, which was trained on raw motion, detects no content change from repair.

\begin{table}[!htbp]
\caption{All BLEU orders for Raw, Global smoothing, and Fused repair on the same 75-sequence content cohort.}
\label{tab:bleu_orders}
\centering
\small
\begin{tabular*}{\linewidth}{l@{\extracolsep{\fill}}rrrr}
\toprule
Human motion & BLEU-1 & BLEU-2 & BLEU-3 & BLEU-4 \\
\midrule
Raw & 54.39 & 41.46 & 32.88 & 26.59 \\
Global smoothing & 48.65 & 37.33 & 30.06 & 24.50 \\
Fused repair & 54.10 & 41.17 & 32.64 & 26.29 \\
\bottomrule
\end{tabular*}
\end{table}

\begin{table}[!htbp]
\caption{Paired content effects across all BLEU orders. The final column adjusts for four orders within each contrast.}
\label{tab:bleu_effects}
\centering
\small
\setlength{\tabcolsep}{4pt}
\begin{tabular*}{\linewidth}{l@{\extracolsep{\fill}}cc}
\toprule
Contrast / metric & Difference [95\% interval] & Four-order interval \\
\midrule
Global$-$Raw / BLEU-1 & $-5.735\;[-8.869, -2.633]$ & $[-9.766, -1.736]$ \\
Global$-$Raw / BLEU-2 & $-4.129\;[-7.056, -1.249]$ & $[-7.973, -0.441]$ \\
Global$-$Raw / BLEU-3 & $-2.825\;[-5.592, -0.080]$ & $[-6.460, 0.656]$ \\
Global$-$Raw / BLEU-4 & $-2.092\;[-4.713, 0.532]$ & $[-5.430, 1.167]$ \\
\midrule
Fused$-$Raw / BLEU-1 & $-0.290\;[-1.420, 0.911]$ & $[-1.782, 1.240]$ \\
Fused$-$Raw / BLEU-2 & $-0.290\;[-1.781, 1.199]$ & $[-2.254, 1.704]$ \\
Fused$-$Raw / BLEU-3 & $-0.248\;[-1.917, 1.399]$ & $[-2.449, 1.900]$ \\
Fused$-$Raw / BLEU-4 & $-0.294\;[-2.014, 1.341]$ & $[-2.581, 1.825]$ \\
\midrule
Fused$-$Global / BLEU-1 & $5.445\;[2.361, 8.645]$ & $[1.583, 9.440]$ \\
Fused$-$Global / BLEU-2 & $3.839\;[0.968, 6.804]$ & $[0.249, 7.690]$ \\
Fused$-$Global / BLEU-3 & $2.576\;[-0.194, 5.493]$ & $[-0.963, 6.289]$ \\
Fused$-$Global / BLEU-4 & $1.798\;[-1.002, 4.727]$ & $[-1.672, 5.517]$ \\
\bottomrule
\end{tabular*}
\end{table}

\FloatBarrier
\subsection{Geometry metrics and certification}
\label{app:geometry}
\label{app:geometry_metrics}

Geometry uses the frozen robot model and the MuJoCo 3.6.0 signed-distance query over all 1,519 pairs of $\mathcal{K}_{\mathrm{eval}}$ (Appendix~\ref{app:platform}). A sample penetrates when its worst-pair depth exceeds $\epsilon=10^{-6}$\,m; depth is reported in millimeters and is zero for separated geometry. Grid and interior samples keep separate denominators, and each quantity in Equation~\ref{eq:geometry_metrics_main} is computed per sequence before averaging. Raw, Global, and Fused motions pass through the same retargeting implementation, joint postprocessing, and finger coupling, so the three Direct conditions differ only in their human input. Fused + Geometry applies both repair stages to the Fused + Direct trajectory. Table~\ref{tab:geometry_effects} gives paired differences from 10,000 sequence-bootstrap resamples (seed 20260916), and Table~\ref{tab:geometry_scope} breaks the Fused result down by pair category.

\begin{table}[!htbp]
\caption{Paired effects for the four-condition geometry comparison over all 1,519 pairs in $\mathcal{K}_{\mathrm{eval}}$ and 20,648 matched sequences. Entries give the difference and 95\% interval; pp denotes percentage points. Geometry denotes Fused + Geometry.}
\label{tab:geometry_effects}
\centering
\footnotesize
\setlength{\tabcolsep}{3pt}
\begin{tabular*}{\linewidth}{l@{\extracolsep{\fill}}ccc}
\toprule
Metric & \makecell{Global$-$Raw\\(Direct)} & \makecell{Geometry$-$\\Fused + Direct} & \makecell{Geometry$-$\\Global + Direct} \\
\midrule
Pen. frames (pp) & \makecell{$-0.637$\\$[-0.706, -0.570]$} & \makecell{$-49.893$\\$[-50.204, -49.576]$} & \makecell{$-52.182$\\$[-52.434, -51.925]$} \\
Inter-frame pen. (pp) & \makecell{$-0.663$\\$[-0.730, -0.597]$} & \makecell{$-49.902$\\$[-50.213, -49.585]$} & \makecell{$-52.195$\\$[-52.448, -51.936]$} \\
Mean pen. depth (mm) & \makecell{$-0.383$\\$[-0.419, -0.347]$} & \makecell{$-8.270$\\$[-8.357, -8.184]$} & \makecell{$-10.991$\\$[-11.081, -10.903]$} \\
Pen.-free clips (pp) & \makecell{$0.140$\\$[0.082, 0.199]$} & \makecell{$87.180$\\$[86.720, 87.631]$} & \makecell{$88.178$\\$[87.742, 88.624]$} \\
\bottomrule
\end{tabular*}
\end{table}

\begin{table}[!htbp]
\caption{Per-category penetration for Fused + Direct and Fused + Geometry ($\mathcal{G}^{\mathrm{repair}}$) at 50\,Hz across all 20,648 sequences. All entries are macro percentages of penetrating samples. Interior samples use three fixed points per resampled interval.}
\label{tab:geometry_scope}
\centering
\small
\setlength{\tabcolsep}{3pt}
\begin{tabular*}{\linewidth}{l@{\extracolsep{\fill}}rrrr}
\toprule
& \multicolumn{2}{c}{Grid frames (\%)} & \multicolumn{2}{c}{Interior samples (\%)} \\
\cmidrule(lr){2-3}\cmidrule(lr){4-5}
Pair scope & Direct & Geometry ($\mathcal{G}^{\mathrm{repair}}$) & Direct & Geometry ($\mathcal{G}^{\mathrm{repair}}$) \\
\midrule
Inter-hand + hand-body & 26.6152 & 0.0993 & 26.6520 & 0.1140 \\
Intra-hand & 31.4145 & 0.2841 & 31.4028 & 0.2848 \\
All audited pairs & 50.1835 & 0.2904 & 50.2044 & 0.3027 \\
\bottomrule
\end{tabular*}
\par\vspace{2pt}
\raggedright\footnotesize
\textit{Note:} Across all audited pairs, $\mathcal{G}^{\mathrm{repair}}$ lowers penetrating samples by 99.4\% relative to Direct on both the grid and interior samples. On the 18,214 certified sequences, no sample penetrates in any scope.
\end{table}

On the 50\,Hz samples, 18,251 sequences (88.39\%) are penetration-free. Certification additionally requires clearance on the 30\,Hz grid and holds for 18,214 sequences (88.21\% of $\mathcal{G}^{\mathrm{repair}}$). Table~\ref{tab:geometry_construction} summarizes construction coverage, certification, deformation, and runtime. Direct fails the temporal limits in every sequence. Joint deviation is the RMS change over the 14 arm coordinates and grid frames, and relation change is the RMS change of the right-minus-left wrist vector in the torso frame. The repaired means are 0.1886\,rad (95\% interval $[0.1869,0.1904]$, about $10.8^\circ$) and 7.955\,cm ($[7.877,8.036]$). The relation change follows from the robot's body: its forearm and wrist links have a 4--5\,cm collision radius, so forearm crossings that nearly touch on a human signer must be opened, while $\mathcal{E}_{\mathrm{rel}}$ keeps the wrist vector otherwise close to the human reference. The runtime is concurrent CPU wall-clock time per sequence, with medians of 2.25 and 179.97\,s; geometry repair adds 364.48\,s on average (95\% interval $[358.42,370.64]$).

\begin{table}[!htbp]
\caption{Construction outcome for the Fused production pair. Counts use all 20,652 planned inputs; deformation and runtime average over the 20,648 completed pairs. RMS measures the change from Direct, and runtime is concurrent CPU wall-clock time per sequence.}
\label{tab:geometry_construction}
\label{tab:geometry_acceptance}
\centering
\small
\setlength{\tabcolsep}{3pt}
\begin{tabular*}{\linewidth}{l@{\extracolsep{\fill}}rrrccc}
\toprule
Robot reference & Output & \makecell{Checks\\passed} & Certified & \makecell{Joint RMS\\(rad)} & \makecell{Relation RMS\\(cm)} & \makecell{Runtime\\(s)} \\
\midrule
Fused + Direct & 20,649 & 0 & 0 & 0.0000 & 0.000 & 2.46 \\
Fused + Geometry & 20,648 & 20,648 & 18,214 & 0.1886 & 7.955 & 366.93 \\
\bottomrule
\end{tabular*}
\end{table}

\label{app:intrahand_results}
\paragraph{Sample-level clearance.}
Over all 16,234,504 grid and interior samples, 0.31\% of $\mathcal{G}^{\mathrm{repair}}$ samples penetrate, compared with 50.54\% for Direct; the sequence averages in Table~\ref{tab:geometry_scope} are 0.29\% and 50.18\% on the grid. In the 18,214 certified sequences, no sample penetrates on either grid, and the largest recorded depth is 0.000997\,mm. The finger stage changes the active joints by a frame-weighted RMS of $1.40^\circ$.

\paragraph{Handshape deformation budget.}
The per-joint budget $\theta_{\max}$ bounds how far the finger stage may change a handshape. On sequence \texttt{S000374\_P0004\_T00} (Table~\ref{tab:control_semantics}), budgets of $5^\circ$ and $10^\circ$ leave intra-hand overlaps of 14.2 and 13.8\,mm, whereas $20^\circ$ clears the sequence with a largest active-joint change of $11.7^\circ$. We use $20^\circ$ for the full corpus. Across the 15,714 certified sequences that required finger changes, the largest active-joint change per sequence has a median of $8.6^\circ$ and a 90th percentile of $16.8^\circ$, and 461 sequences reach the budget. Because the objective minimizes displacement, the frame-weighted RMS stays at $1.40^\circ$. The budget applies to the active joints, and the passive joints follow them through the mechanical coupling.

\paragraph{Uncertified sequences.}
Table~\ref{tab:uncertified} divides the 2,434 uncertified sequences by the check they fail; each sequence belongs to one group.
(i)~\emph{Control-grid residuals} (762 sequences) pass every 30\,Hz check and fail only on the 50\,Hz grid, which the finger stage does not sample.
(ii)~\emph{Intra-hand residuals} (615) keep intra-hand penetration at 30\,Hz without contact between parts; 562 of them end at the $20^\circ$ budget on at least one active joint (within $10^{-3}$\,deg).
(iii)~\emph{Cross-part residuals} (1,057) fail an inter-hand or hand-body check at 30\,Hz, usually together with an intra-hand residual; 810 end at the budget.
The 2,434 uncertified sequences retain contact on 2.2\% of their samples, and 1,424 of them end at the budget. Their mean worst-pair depth, averaged over all their samples including separated ones, is 0.037\,mm. Per-pair residual records are released with each sequence.

\begin{table}[!htbp]
\caption{Partition of the 2,434 uncertified sequences by failed validation check. Groups are mutually exclusive. Penetrating samples pool the 50\,Hz grid and interior samples over all audited pairs.}
\label{tab:uncertified}
\centering
\small
\setlength{\tabcolsep}{6pt}
\begin{tabular*}{\linewidth}{l@{\extracolsep{\fill}}rrr}
\toprule
Group & Sequences & At $20^\circ$ cap & \makecell{Penetrating\\samples (\%)} \\
\midrule
(i) Control-grid only & 762 & 52 & 0.18 \\
(ii) Native intra-hand & 615 & 562 & 3.03 \\
(iii) Native cross-part & 1,057 & 810 & 3.19 \\
\midrule
All uncertified & 2,434 & 1,424 & 2.21 \\
\bottomrule
\end{tabular*}
\end{table}

\FloatBarrier
\subsection{Execution metrics and replay protocol}
\label{app:control_metrics}

Table~\ref{tab:tolerance} lists the two tolerance profiles. The nominal profile defines the pass thresholds and the ActionScore scales of Section~\ref{sec:execution_score}; the alternative profile is used only for sensitivity analysis (Appendix~\ref{app:threshold_auc}). A component passes when $e_t^x\leq\sigma_x$. Handshape, location, and orientation use the worse hand or the largest active-joint error, and relation uses the right-minus-left wrist vector.

\begin{table}[!htbp]
\caption{Nominal and alternative tolerance profiles. The palm tolerances are about $28.6^\circ$ and $45.8^\circ$.}
\label{tab:tolerance}
\centering
\small
\setlength{\tabcolsep}{4pt}
\begin{tabularx}{\linewidth}{@{}lXrr@{}}
\toprule
Component & Per-frame error & Nominal & Alternative \\
\midrule
Handshape & Largest absolute error over 12 active joints & 0.30\,rad & 0.50\,rad \\
Location & Larger left/right wrist position error & 10\,cm & 15\,cm \\
Orientation & Larger left/right palm rotation error & $0.50$\,rad & $0.80$\,rad \\
Relation & Error in the right-minus-left wrist vector & 10\,cm & 15\,cm \\
\bottomrule
\end{tabularx}
\end{table}

Component attainment (CA) is the per-component pass rate over all planned frames,
\begin{equation}
\label{eq:component_attainment}
\operatorname{CA}_{i,s}=\frac{100}{4T_i}\sum_{t=0}^{T_i-1}A_{i,s,t}
\sum_{x\in\{\mathrm{hand},\mathrm{pos},\mathrm{ori},\mathrm{rel}\}}\mathbf{1}[e_t^x\leq\sigma_x],
\end{equation}
so unreached frames count as failures. Table~\ref{tab:hard_components} reports the pass rate of each component.

\paragraph{Worked example.}
Consider 100 planned frames, of which 80 are reached, and normalized errors of 0, 0.5, 1, and 2 for the four components on every reached frame. The per-frame fidelities are 1, 0.8409, 0.5, and 0.0625, giving component scores of 80, 67.27, 40, and 5 and an ActionScore of 48.07. Three components pass on each reached frame, so CA is 60\%; JointPass is zero because relation fails throughout. Coverage is 80\%, and the trial is unfinished.

\paragraph{Replay.}
Each replay tracks the complete reference, including the full finger target, from the first frame on the 50\,Hz grid, with one recorded state per reference frame and no temporal alignment. The references contain only the signing motion, so every scored frame is a signing frame. Scores use the measured simulation state rather than the commanded target. The torso frame $B$ is the torso link; wrist positions are taken at the wrist-yaw links and palm orientations at the palm links. The active coordinates of each hand are the two thumb joints and the proximal joints of the index, middle, ring, and little fingers. A trial ends at its first failure: anchor-height error above 0.25\,m, a projected-gravity $z$ discrepancy above 0.8, an ankle-height error above 0.25\,m, or a numerical failure. Ordinary episode timeouts are disabled, so the reference length determines completion. Tracking metrics exclude the failing frame and all later frames, whereas penetration uses every recorded state, including the failing one. Motion after a reset is never appended to a terminated rollout.

For motion $i$ with frame index $\mathcal{I}_i$, coverage and the reached-frame RMS of component $x$ are
\begin{equation}
\label{eq:coverage_rms}
\operatorname{Coverage}_{i,s}=\frac{\sum_{t\in\mathcal{I}_i}A_{i,s,t}}{|\mathcal{I}_i|},
\qquad
\operatorname{RMS}^{x}_{\mathrm{reached}}=
\sqrt{\frac{\sum_{t\in\mathcal{I}_i:A_{i,s,t}=1}(e_t^x)^2}
{\sum_{t\in\mathcal{I}_i}A_{i,s,t}}}.
\end{equation}
An empty reached set gives no value. Hand RMS averages the squared error over the twelve active joints, whereas the handshape pass test uses their largest error. The physical indicator $G_{i,s,t}$ of Equation~\ref{eq:joint_pass} uses the same $10^{-6}$\,m tolerance over $\mathcal{K}_{\mathrm{eval}}$ as rollout penetration. Rollout penetration counts a recorded frame once if any audited pair penetrates, and unexecuted frames are not counted as collision-free. We also report the share of recorded frames deeper than 1\,mm and the mean and peak worst-pair depth, with zero for separated frames. The coupling error (Table~\ref{tab:control_diagnostics}) is the RMS deviation of the executed passive finger joints from the coupling applied to the executed active joints, with each coupled value clipped to its soft joint range.

For active finger joint $j$ mapped to human hand group $g(j)$, the high-support subset of finger targets is
\begin{equation}
\label{eq:provenance_subset}
\mathcal{T}_{\mathrm{prov}}
=\{(t,j)\in\mathcal{I}_i\times\mathcal{J}_{\mathrm{hand}}\mid c_t^{g(j)}>0.5\},
\end{equation}
where $c_t^g$ is the score of Table~\ref{tab:provenance}. It is used only for the diagnostics in Appendix~\ref{app:provenance_stratified}; all benchmark scores use the full index.

\paragraph{Temporal tracking.}
Velocity and acceleration errors compare executed and reference changes on the 50\,Hz grid. For the active hand joints, the torso-relative wrists, and the inter-wrist vector,
\begin{equation}
\label{eq:temporal_tracking}
\delta v_t=\frac{(z^{\mathrm{exec}}_{t+1}-z^{\mathrm{exec}}_t)-(z^{\mathrm{ref}}_{t+1}-z^{\mathrm{ref}}_t)}{\Delta t},\qquad
\delta a_t=\frac{\delta v_{t+1}-\delta v_t}{\Delta t},\qquad \Delta t=0.02\,\mathrm{s}.
\end{equation}
For palms, the angular velocity is $\omega_t=\operatorname{Log}(\widetilde R_{t+1}\widetilde R_t^\top)^\vee/\Delta t$, and its residual is differenced to give the angular-acceleration error. No smoothing or temporal warping is applied. A velocity window requires both frames to be reached and an acceleration window all three; window coverage uses the planned counts $T_i-1$ and $T_i-2$. A constant position offset has zero velocity error, so these metrics complement the position errors.

\FloatBarrier
\subsection{Aggregation and failure accounting}
\label{app:statistics}

The sequence or motion is the unit of inference. Human and geometry quantities are summarized per sequence. For control, our controller has 480 self-target replays (20 motions, seeds 42, 1234, and 777, eight configurations), and AMP has 180 (three reference tracks). Each motion receives equal weight after averaging its seeds. Paired intervals resample the 20 motions as blocks, keeping all conditions and seeds of a motion together, so they describe variation across motions given the three observed seeds. AMP reference-track effects use the same procedure. AMP and our controller are trained differently, so their comparison is reported without intervals; every AMP replay stops early, whereas our controller completes every replay on the repaired reference. Human and geometry intervals resample sequences.

Every result distinguishes planned, input-valid, generated, trained, and evaluated counts. When a task reference exists, a failure of reference generation, training, or evaluation receives zero ActionScore and counts as unfinished; its conditional errors are reported as missing. Interrupted runs resume under the same run identifier. Numerical instability, repeated falls, and non-convergence are kept as outcomes and are not replaced by another seed or checkpoint.

\FloatBarrier
\section{Supplementary results}
\label{app:additional}

\FloatBarrier
\subsection{Complete control results}
\label{app:seeds}

Table~\ref{tab:control} gives the complete results behind Table~\ref{tab:control_main}, adding Global smoothing, the no-ramp curriculum, JointPass, and penetration depth. Table~\ref{tab:control_seeds} reports each seed. Tables~\ref{tab:control_tracking}--\ref{tab:control_velocity} decompose the same trials into coverage, reached-frame errors, physical diagnostics, component scores, pass rates, and temporal errors; they are views of the same replays, not independent trials. All intervals below use 10,000 paired bootstrap resamples of the 20 motions with the three seeds kept within each motion, and all differences are computed from unrounded values.

\paragraph{Reference processing.}
With single-stage training, Fused raises finished replays from 34/60 for Raw to 58/60, a paired gain of 40.00 percentage points (95\% interval $[23.33,56.67]$). Raw policies stop earlier, so their reached-frame errors cover shorter intervals (Table~\ref{tab:control_tracking}). Global smoothing also makes references learnable (ActionScore 50.17), but it lowers BLEU-1 and BLEU-2 relative to Raw (Section~\ref{sec:human_results}), whereas Fused keeps the recognized content; adding geometry repair to Fused gives the highest scores. Under the two-stage curriculum, geometry repair lowers the share of frames deeper than 1\,mm from 10.44\% to 4.31\% (paired change $-6.13$ percentage points, $[-12.56,-1.29]$) and hand RMS from 0.0878 to 0.0761\,rad.

\paragraph{Curriculum.}
On the fixed Fused + Geometry reference, all three curricula finish all 60 replays. The full curriculum raises ActionScore by 1.52 over single-stage training and by 0.66 over no-ramp training, and it is higher for each of the three seeds (Table~\ref{tab:control_seeds}). Relative to single-stage training, it raises the pass rate of all four components (Table~\ref{tab:hard_components}). Relative to no-ramp training, it lowers palm RMS by $0.76^\circ$ ($[-1.49,-0.06]$) and raises all four continuous component scores (Table~\ref{tab:score_components}). Both two-stage curricula also lower the hand, wrist, and palm velocity and acceleration errors relative to single-stage training (Table~\ref{tab:control_velocity}); with and without the ramp, these temporal errors are similar.

\paragraph{AMP.}
Every AMP replay stops after 10--12 recorded frames, a coverage of 5.27--5.28\% on all three reference tracks. The paired ActionScore effects are $-0.058$ for Fused + Direct minus Raw + Direct ($[-0.346,0.185]$), $+0.117$ for Fused + Geometry minus Fused + Direct ($[-0.014,0.255]$), and $+0.059$ for Fused + Geometry minus Raw + Direct ($[-0.180,0.274]$).

\begin{table}[!htbp]
\centering
\caption{Complete controlled execution results on 20 motions and three seeds. Within each group, rows differ only in the reference; the Fused + Geometry rows of (a)--(c) differ only in the curriculum. Shading marks our full pipeline, and bold marks the best value among our controllers.}
\label{tab:control}
\footnotesize
\setlength{\tabcolsep}{3.5pt}
\renewcommand{\arraystretch}{1.04}
\begin{tabularx}{\linewidth}{@{}l*{5}{>{\centering\arraybackslash}X}@{}}
\toprule
Controller / reference & ActionScore $\uparrow$ & JointPass (\%) $\uparrow$ & Finished (\%) $\uparrow$ & Pen.\ (\%) $\downarrow$ & Depth (mm) $\downarrow$ \\
\midrule
\multicolumn{6}{@{}l}{\textbf{(a)}\enspace\textit{Ours, single-stage}} \\
Raw + Direct & 34.99 & 7.50 & 56.7 & 36.15 & 0.548 \\
Global + Direct & 50.17 & 12.46 & 100.0 & 27.67 & 0.309 \\
Fused + Direct & 49.15 & 13.56 & 96.7 & 25.97 & 0.865 \\
Fused + Geometry & 52.58 & 16.32 & 100.0 & 16.09 & 0.138 \\
\midrule
\multicolumn{6}{@{}l}{\textbf{(b)}\enspace\textit{Ours, two-stage + ramp}} \\
Raw + Direct & 35.51 & 7.03 & 60.0 & 35.53 & 0.539 \\
Fused + Direct & 49.29 & 12.25 & 98.3 & 26.08 & 0.663 \\
\rowcolor{tableaccent} \textbf{Fused + Geometry} & \textbf{54.10} & \textbf{17.98} & \textbf{100.0} & \textbf{13.03} & 0.112 \\
\midrule
\multicolumn{6}{@{}l}{\textbf{(c)}\enspace\textit{Ours, two-stage without ramp}} \\
Fused + Geometry & 53.44 & 17.10 & 100.0 & 13.55 & \textbf{0.104} \\
\midrule
\multicolumn{6}{@{}l}{\textbf{(d)}\enspace\textit{AMP, native multi-motion training}} \\
Raw + Direct & 2.73 & 0.71 & 0.0 & 32.24 & 1.604 \\
Fused + Direct & 2.68 & 0.49 & 0.0 & 26.39 & 3.038 \\
Fused + Geometry & 2.79 & 0.96 & 0.0 & 23.50 & 0.475 \\
\bottomrule
\end{tabularx}
\par\smallskip
\begin{minipage}{\linewidth}\footnotesize
Unreached frames score zero. Pen.\ and Depth use recorded states and all audited pairs, including intra-hand pairs.
\end{minipage}
\end{table}

\begin{table}[!htbp]
\caption{Per-seed control results on the same 20 motions. The main table averages seeds within each motion before averaging motions. Finished measures replay completion; ActionScore also accounts for reached-frame fidelity.}
\label{tab:control_seeds}
\centering\small
\setlength{\tabcolsep}{3pt}
\begin{tabularx}{\linewidth}{@{}l*{6}{>{\centering\arraybackslash}X}@{}}
\toprule
& \multicolumn{3}{c}{ActionScore $\uparrow$} & \multicolumn{3}{c}{Finished (\%) $\uparrow$} \\
\cmidrule(lr){2-4}\cmidrule(lr){5-7}
Reference & 42 & 1234 & 777 & 42 & 1234 & 777 \\
\midrule
\multicolumn{7}{@{}l}{\textit{(a) Single-stage training}} \\
Raw + Direct & 30.00 & 36.93 & 38.05 & 45 & 60 & 65 \\
Global + Direct & 49.27 & 50.30 & 50.92 & 100 & 100 & 100 \\
Fused + Direct & 50.13 & 49.46 & 47.85 & 100 & 100 & 90 \\
Fused + Geometry & 52.20 & 52.03 & 53.51 & 100 & 100 & 100 \\
\midrule
\multicolumn{7}{@{}l}{\textit{(b) Two-stage training}} \\
Raw + Direct & 34.82 & 36.00 & 35.70 & 65 & 60 & 55 \\
Fused + Direct & 49.36 & 49.84 & 48.68 & 95 & 100 & 100 \\
Fused + Geometry, no ramp & 53.46 & 53.27 & 53.60 & 100 & 100 & 100 \\
Fused + Geometry & 54.05 & 53.74 & 54.51 & 100 & 100 & 100 \\
\bottomrule
\end{tabularx}
\end{table}

\begin{table}[!htbp]
\caption{Reached-frame tracking errors and deeper penetration for the same trials as Table~\ref{tab:control}. Coverage exposes the interval on which conditional errors are measured. The fixed Fused + Geometry curriculum comparison has complete coverage in every condition.}
\label{tab:control_tracking}
\centering\small
\setlength{\tabcolsep}{3pt}
\renewcommand{\arraystretch}{1.14}
\begin{tabularx}{\linewidth}{@{}l*{5}{>{\centering\arraybackslash}X}@{}}
\toprule
& Coverage & \multicolumn{3}{c}{Reached-frame RMS $\downarrow$} & Pen. $>1$\,mm \\
\cmidrule(lr){2-2}\cmidrule(lr){3-5}\cmidrule(lr){6-6}
Reference & (\%) $\uparrow$ & Hand (rad) & Wrist (cm) & Palm ($^\circ$) & (\%) $\downarrow$ \\
\midrule
\multicolumn{6}{@{}l}{\textit{(a) Single-stage training: reference processing}} \\
Raw + Direct & 68.8 & 0.0896 & 13.02 & 48.50 & 15.69 \\
Global + Direct & 100.0 & 0.0821 & 14.45 & 43.65 & 13.08 \\
Fused + Direct & 98.4 & 0.0926 & 14.55 & 43.81 & 10.51 \\
Fused + Geometry & 100.0 & 0.0792 & 13.97 & 40.97 & 5.51 \\
\midrule
\multicolumn{6}{@{}l}{\textit{(b) Two-stage training: reference processing and ramp ablation}} \\
Raw + Direct & 72.1 & 0.0859 & 13.47 & 48.20 & 15.74 \\
Fused + Direct & 98.7 & 0.0878 & 14.44 & 43.62 & 10.44 \\
Fused + Geometry, no ramp & 100.0 & 0.0770 & 13.59 & 39.96 & 3.94 \\
\rowcolor{tableaccent} \textbf{Fused + Geometry} & 100.0 & 0.0761 & 13.42 & 39.19 & 4.31 \\
\bottomrule
\end{tabularx}
\par\smallskip
\begin{minipage}{\linewidth}\footnotesize
Hand RMS is the square root of the mean squared error over the twelve active finger joints and reached frames. Wrist and palm RMS use the larger bilateral error per frame. All RMS errors use reached frames before the first failure; no unexecuted suffix is imputed. The $>1$\,mm rate uses all recorded physical frames. These conditional diagnostics supplement the planned-interval action scores in the main table.
\end{minipage}
\end{table}

\begin{table}[!htbp]
\caption{Replay stability and physical diagnostics for the same trials as Table~\ref{tab:control}. Means average seeds within each motion and give equal weight to each motion. Depth statistics use the positive depth of the worst audited pair at each executed frame, with zero for separated pairs.}
\label{tab:control_diagnostics}
\centering\small
\setlength{\tabcolsep}{2pt}
\renewcommand{\arraystretch}{1.14}
\begin{tabularx}{\linewidth}{@{}l*{7}{>{\centering\arraybackslash}X}@{}}
\toprule
& \multicolumn{2}{c}{Replay} & \multicolumn{3}{c}{Additional diagnostics $\downarrow$} & \multicolumn{2}{c}{Depth (mm) $\downarrow$} \\
\cmidrule(lr){2-3}\cmidrule(lr){4-6}\cmidrule(lr){7-8}
Reference & \makecell{Finished\\(\%) $\uparrow$} & \makecell{Coverage\\(\%) $\uparrow$} & \makecell{Relation\\(cm)} & \makecell{Limit\\(\%)} & \makecell{Coupling\\(rad)} & Mean & Peak \\
\midrule
\multicolumn{8}{@{}l}{\textit{(a) Single-stage training: reference processing}} \\
Raw + Direct & 56.7 & 68.8 & 13.37 & 0.476 & 0.0957 & 0.548 & 9.21 \\
Global + Direct & 100.0 & 100.0 & 12.32 & 0.274 & 0.0629 & 0.309 & 6.33 \\
Fused + Direct & 96.7 & 98.4 & 11.37 & 0.307 & 0.0687 & 0.865 & 10.65 \\
Fused + Geometry & 100.0 & 100.0 & 10.38 & 0.498 & 0.0614 & 0.138 & 3.29 \\
\midrule
\multicolumn{8}{@{}l}{\textit{(b) Two-stage training: reference processing and ramp ablation}} \\
Raw + Direct & 60.0 & 72.1 & 13.65 & 0.383 & 0.0936 & 0.539 & 9.00 \\
Fused + Direct & 98.3 & 98.7 & 11.05 & 0.323 & 0.0667 & 0.663 & 10.62 \\
Fused + Geometry, no ramp & 100.0 & 100.0 & 10.11 & 0.348 & 0.0613 & 0.104 & 3.30 \\
\rowcolor{tableaccent} \textbf{Fused + Geometry} & 100.0 & 100.0 & 10.05 & 0.188 & 0.0612 & 0.112 & 3.36 \\
\bottomrule
\end{tabularx}
\par\smallskip
\begin{minipage}{\linewidth}\footnotesize
Finished is the percentage of complete replays without early termination; it does not test action accuracy. Coverage is reached/planned frames. Nominal JointPass is reported in Table~\ref{tab:control}. Limit violations use a $10^{-3}$\,rad allowance over recorded physical frames. Relation and coupling are RMS errors over reached frames. Mean depth includes all recorded frames, rather than only penetrating frames. Peak depth is the macro-average of per-trial maximum depths across seeds and motions. Appendix~\ref{app:control_metrics} defines the common protocol.
\end{minipage}
\end{table}

\begin{table}[!htbp]
\caption{ActionScore decomposition and tolerance sensitivity. The first four columns report component scores under the nominal scales using the complete planned-frame denominator; their mean gives the nominal ActionScore. Alternative-scale ActionScore applies the alternative tolerance profile to the same recorded trajectories. CA reports component-wise hard-threshold attainment under the nominal profile. All columns are higher-is-better.}
\label{tab:score_components}
\centering\small
\setlength{\tabcolsep}{2.4pt}
\renewcommand{\arraystretch}{1.14}
\begin{tabularx}{\linewidth}{@{}l*{7}{>{\centering\arraybackslash}X}@{}}
\toprule
& \multicolumn{4}{c}{Continuous component score} & \multicolumn{2}{c}{ActionScore} & CA \\
\cmidrule(lr){2-5}\cmidrule(lr){6-7}\cmidrule(lr){8-8}
Reference & \makecell{Hand-\\shape} & Location & Orient. & Relation & Nominal & Alt. & \makecell{CA\\nom. (\%)} \\
\midrule
\multicolumn{8}{@{}l}{\textit{(a) Single-stage training: reference processing}} \\
Raw + Direct & 53.55 & 30.56 & 22.80 & 33.05 & 34.99 & 47.84 & 35.94 \\
Global + Direct & 81.24 & 36.11 & 35.30 & 48.03 & 50.17 & 69.32 & 50.06 \\
Fused + Direct & 77.30 & 34.50 & 33.02 & 51.77 & 49.15 & 68.05 & 48.44 \\
Fused + Geometry & 81.34 & 36.37 & 36.48 & 56.13 & 52.58 & 71.75 & 51.78 \\
\midrule
\multicolumn{8}{@{}l}{\textit{(b) Two-stage training: reference processing and ramp ablation}} \\
Raw + Direct & 57.01 & 29.97 & 22.03 & 33.02 & 35.51 & 49.24 & 35.72 \\
Fused + Direct & 78.47 & 33.83 & 32.77 & 52.09 & 49.29 & 68.50 & 48.21 \\
Fused + Geometry, no ramp & 82.09 & 37.13 & 37.29 & 57.25 & 53.44 & 72.52 & 52.55 \\
\rowcolor{tableaccent} \textbf{Fused + Geometry} & 82.51 & 37.86 & 38.44 & 57.59 & 54.10 & 73.11 & 53.11 \\
\bottomrule
\end{tabularx}
\end{table}

\begin{table}[!htbp]
\caption{Hard-threshold execution metrics under the nominal and alternative tolerance profiles. The first four columns report component-wise attainment over all planned frames under the nominal profile, and their average gives nominal CA. Alternative CA and JointPass use the alternative profile. These quantities are percentages and are distinct from the continuous ActionScore reported in Table~\ref{tab:control}.}
\label{tab:hard_components}
\centering\small
\setlength{\tabcolsep}{2.4pt}
\begin{tabularx}{\linewidth}{@{}l*{7}{>{\centering\arraybackslash}X}@{}}
\toprule
Reference & \makecell{Hand-\\shape} & Location & Orient. & Relation &
\makecell{CA\\nominal} &
\makecell{CA\\alternative} &
\makecell{JointPass\\alternative} \\
\midrule
\multicolumn{8}{@{}l}{\textit{(a) Single-stage training: reference processing}} \\
Raw + Direct & 61.26 & 29.67 & 18.30 & 34.53 & 35.94 & 54.12 & 23.47 \\
Global + Direct & 92.62 & 29.89 & 29.66 & 48.07 & 50.06 & 76.50 & 37.67 \\
Fused + Direct & 86.40 & 27.77 & 25.63 & 53.94 & 48.44 & 75.47 & 41.26 \\
Fused + Geometry & 91.05 & 28.89 & 28.96 & 58.21 & 51.78 & 80.35 & 49.95 \\
\midrule
\multicolumn{8}{@{}l}{\textit{(b) Two-stage training: reference processing and ramp ablation}} \\
Raw + Direct & 65.17 & 27.64 & 16.48 & 33.60 & 35.72 & 55.03 & 23.16 \\
Fused + Direct & 87.89 & 26.65 & 24.91 & 53.37 & 48.21 & 75.97 & 40.75 \\
Fused + Geometry, no ramp & 91.12 & 28.84 & 29.72 & 60.52 & 52.55 & 81.82 & 52.89 \\
\rowcolor{tableaccent} \textbf{Fused + Geometry} & 91.68 & 29.88 & 31.21 & 59.68 & 53.11 & 82.88 & 53.67 \\
\bottomrule
\end{tabularx}
\end{table}

\begin{table}[!htbp]
\caption{Velocity and acceleration tracking errors relative to the reference. RMS is computed on reached windows per trial, averaged across seeds within each motion and then equally across motions. Window coverage uses the planned denominator. All error columns are lower-is-better.}
\label{tab:control_velocity}
\label{tab:control_acceleration}
\centering\small
\setlength{\tabcolsep}{2.4pt}
\begin{tabularx}{\linewidth}{@{}l*{5}{>{\centering\arraybackslash}X}@{}}
\toprule
Reference & \makecell{Windows\\(\%) $\uparrow$} & Hand & Wrist & Palm & Relation \\
\midrule
\multicolumn{6}{@{}l}{\textbf{Velocity error} (hand and palm in rad/s; wrist and relation in m/s)} \\
\multicolumn{6}{@{}l}{\textit{(a) Single-stage training: reference processing}} \\
Raw + Direct & 68.6 & 1.042 & 1.212 & 11.546 & 1.324 \\
Global + Direct & 100.0 & 0.741 & 0.721 & 5.090 & 0.816 \\
Fused + Direct & 98.4 & 0.657 & 0.592 & 4.489 & 0.637 \\
Fused + Geometry & 100.0 & 0.517 & 0.556 & 4.072 & 0.616 \\
\midrule
\multicolumn{6}{@{}l}{\textit{(b) Two-stage training: reference processing and ramp ablation}} \\
Raw + Direct & 71.9 & 1.016 & 1.266 & 11.385 & 1.373 \\
Fused + Direct & 98.7 & 0.645 & 0.574 & 4.454 & 0.631 \\
Fused + Geometry, no ramp & 100.0 & 0.503 & 0.546 & 4.043 & 0.619 \\
\rowcolor{tableaccent} \textbf{Fused + Geometry} & 100.0 & 0.504 & 0.546 & 4.033 & 0.621 \\
\midrule
\multicolumn{6}{@{}l}{\textbf{Acceleration error} (hand and palm in rad/s$^2$; wrist and relation in m/s$^2$)} \\
\multicolumn{6}{@{}l}{\textit{(a) Single-stage training: reference processing}} \\
Raw + Direct & 68.5 & 40.586 & 49.558 & 546.092 & 51.900 \\
Global + Direct & 100.0 & 20.675 & 10.513 & 82.978 & 11.697 \\
Fused + Direct & 98.4 & 17.866 & 9.832 & 85.577 & 10.135 \\
Fused + Geometry & 100.0 & 13.191 & 7.591 & 59.646 & 8.206 \\
\midrule
\multicolumn{6}{@{}l}{\textit{(b) Two-stage training: reference processing and ramp ablation}} \\
Raw + Direct & 71.8 & 39.310 & 52.192 & 537.963 & 54.236 \\
Fused + Direct & 98.7 & 17.293 & 9.605 & 85.060 & 9.982 \\
Fused + Geometry, no ramp & 100.0 & 12.848 & 7.398 & 58.572 & 8.198 \\
\rowcolor{tableaccent} \textbf{Fused + Geometry} & 100.0 & 12.991 & 7.409 & 58.630 & 8.231 \\
\bottomrule
\end{tabularx}
\end{table}

\subsection{Tolerance sensitivity}
\label{app:threshold_auc}

Under the alternative profile of Table~\ref{tab:tolerance}, the full method reaches an ActionScore of 73.11 and a JointPass of 53.67\%, compared with 54.10 and 17.98\% under the nominal profile. The ordering of our control conditions is the same under the two profiles (Tables~\ref{tab:score_components} and~\ref{tab:hard_components}).

\paragraph{Common threshold scaling.}
To test orderings between the two profiles and beyond, we scale all four nominal thresholds together by $s\in[0.25,3]$ and recompute CA and JointPass from the same recorded states, reach masks, planned-frame denominators, collision test, and joint-limit allowance. The sweep uses all 480 replays of our controller and all 180 AMP replays, averaging seeds within each motion and weighting motions equally. For a percentage-valued score $M(s)$, the normalized area is $\int_{0.25}^{3}M(s)\,ds/2.75$ on a 0--100 scale. We integrate the empirical step functions exactly at their breakpoints and check every pairwise ordering at the union of breakpoints and both endpoints.

\begin{figure}[!htbp]
\centering
\includegraphics[width=\linewidth]{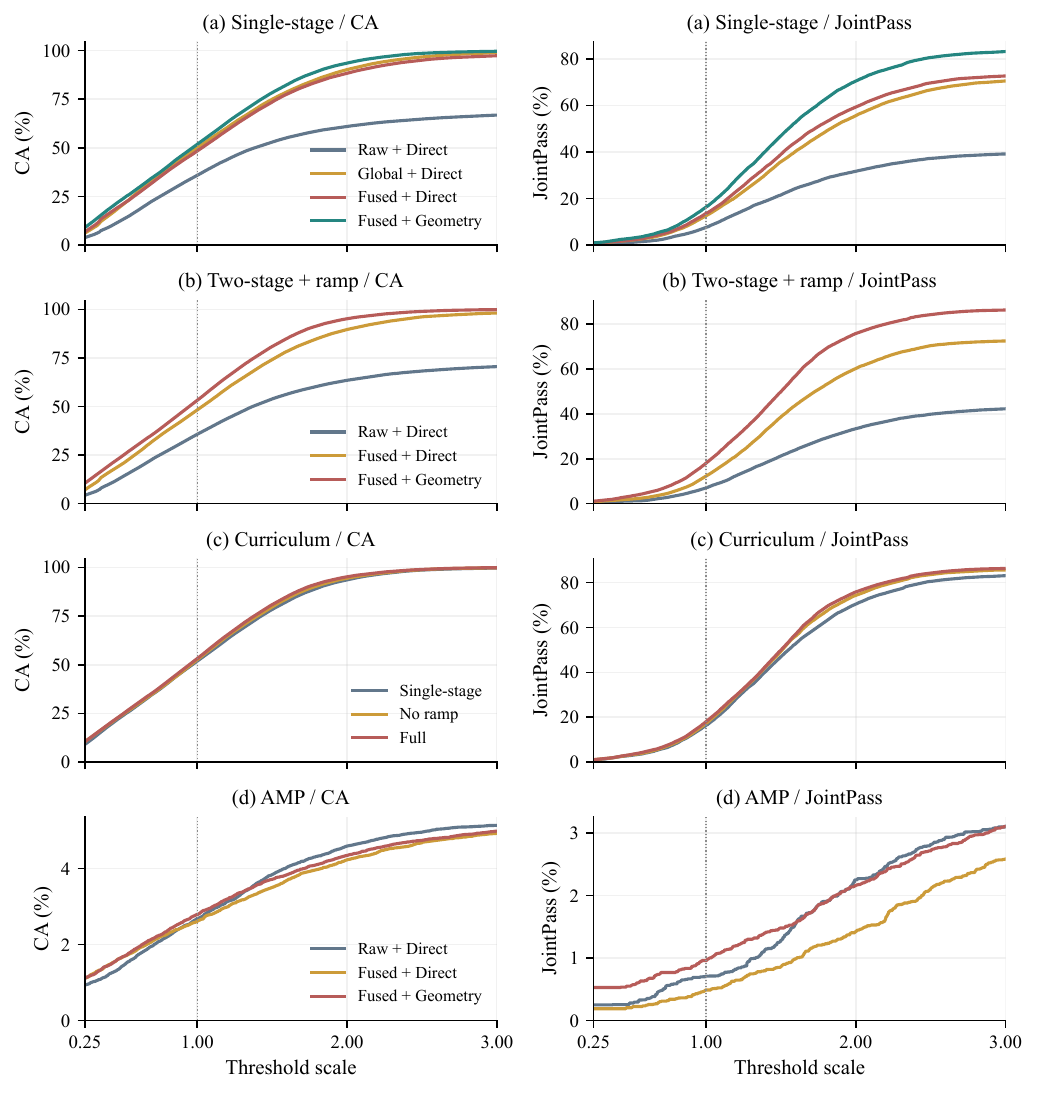}
\caption{CA and JointPass under common scaling of all four action thresholds. Rows (a,b) compare references under single- and two-stage training, (c) compares the three curricula on Fused + Geometry, and (d) shows AMP (Table~\ref{tab:control}). Dotted lines mark the nominal scale. AMP uses a separate vertical scale. Curves are drawn on a dense display grid; areas and ordering checks use the exact empirical breakpoints.}
\label{fig:threshold_auc}
\end{figure}

Fused + Geometry stays above the other references at every scale in panels (a) and (b), for both CA and JointPass (Figure~\ref{fig:threshold_auc}, Table~\ref{tab:threshold_auc}). In panel (c), the full curriculum has the largest CA and JointPass areas. Conditions with close scores exchange order at some scales: Global + Direct and Fused + Direct in panel (a), the three curricula in JointPass in panel (c), and the three AMP reference tracks.

\begin{table}[!htbp]
\centering\small
\caption{Normalized threshold areas on $[0.25,3]$, on a 0--100 scale. Each condition contains 20 motions and three seeds. The panel identifiers match Figure~\ref{fig:threshold_auc}.}
\label{tab:threshold_auc}
\begin{tabular*}{\linewidth}{l@{\extracolsep{\fill}}lrr}
\toprule
Panel & Condition & CA area & JointPass area \\
\midrule
a & Raw + Direct & 47.454 & 21.671 \\
a & Global + Direct & 69.305 & 37.678 \\
a & Fused + Direct & 68.044 & 39.974 \\
a & Fused + Geometry & 71.755 & 46.943 \\
\midrule
b & Raw + Direct & 49.029 & 22.492 \\
b & Fused + Direct & 68.522 & 39.803 \\
b & Fused + Geometry & 73.045 & 49.898 \\
\midrule
c & Single-stage & 71.755 & 46.943 \\
c & Two-stage, no ramp & 72.498 & 49.165 \\
c & Two-stage + ramp & 73.045 & 49.898 \\
\midrule
AMP & Raw + Direct & 3.639 & 1.616 \\
AMP & Fused + Direct & 3.475 & 1.157 \\
AMP & Fused + Geometry & 3.583 & 1.717 \\
\bottomrule
\end{tabular*}
\end{table}

\paragraph{Articulation-site spacing.}
\label{app:articulation_spacing}
On the neutral SMPL-X template, we measure distances between facial landmarks: each eye center averages its six eyelid points, the nose is the nose tip, the mouth center averages the inner-lip midpoints, and the chin is the central contour point. The eye-to-nose distances are 5.8\,cm on both sides, the inter-eye distance is 6.3\,cm, nose-to-mouth is 4.4\,cm, and mouth-to-chin is 4.9\,cm. The 10\,cm location tolerance therefore exceeds the spacing of some neighboring facial sites; it is a declared operating point rather than a bound derived from anatomical spacing.

\subsection{Provenance-stratified tracking error}
\label{app:provenance_stratified}

Because the repair records follow each reference into its rollouts, tracking errors can be grouped by the operation that produced each reference segment. This analysis trains one tracking policy with the architecture of Appendix~\ref{app:control} on the 20 motions of $\mathcal{D}_{\mathrm{control}}$ joined in a fixed order by synthetic transitions, with seeds 42, 1234, and 777, and is replayed over the whole joined reference. Table~\ref{tab:provenance_attribution} reports the joint RMS error of this policy for each operation category and body side, averaged over seeds.

\begin{table}[!htbp]
\caption{Tracking error of the multi-motion policy grouped by reference provenance. Share is the percentage of frames of that side; RMS pools joint-frames and averages the three seeds.}
\label{tab:provenance_attribution}
\centering
\small
\setlength{\tabcolsep}{4pt}
\begin{tabular*}{\linewidth}{l@{\extracolsep{\fill}}rrrrrrr}
\toprule
& & \multicolumn{3}{c}{Right side} & \multicolumn{3}{c}{Left side} \\
\cmidrule(lr){3-5}\cmidrule(lr){6-8}
Operation & Score & \makecell{Share\\(\%)} & \makecell{Arm RMS\\(rad)} & \makecell{Hand RMS\\(rad)} & \makecell{Share\\(\%)} & \makecell{Arm RMS\\(rad)} & \makecell{Hand RMS\\(rad)} \\
\midrule
Retained source fit & 1.00 & 65.43 & 0.162 & 0.062 & 36.08 & 0.154 & 0.057 \\
Exit transition & 0.70 & 0.12 & 0.129 & 0.054 & 2.00 & 0.173 & 0.059 \\
Two-anchor interpolation & 0.50 & 19.07 & 0.179 & 0.070 & 6.23 & 0.181 & 0.070 \\
Single-anchor freezing & 0.25 & 3.52 & 0.150 & 0.105 & 2.16 & 0.148 & 0.057 \\
Long-segment freezing & 0.10 & 0.50 & 0.240 & 0.061 & 42.17 & 0.105 & 0.065 \\
Synthetic transition & -- & 11.36 & 0.153 & 0.044 & 11.36 & 0.113 & 0.044 \\
\bottomrule
\end{tabular*}
\end{table}

The records separate two kinds of held segments. On the right side, which carries most of the signing, long held spans replace motion that the video does not support, and arm error rises from 0.162\,rad RMS on retained fits to 0.240\,rad. On the left side, long held spans cover 42.17\% of frames, since in many control motions the left arm rests for long periods, and such a static pose is tracked easily (0.105\,rad). Without the records, both cases would be pooled with observed motion; with them, a controller developer can separate error on held segments from error on observed motion.

The records can also guide training. Score weighting multiplies each per-joint and per-body squared error in the tracking rewards by the score of its part group, so held and interpolated segments contribute less supervision. Hard rejection instead removes the seven motions whose mean arm score is below 0.5 and trains the same policy on the remaining 13. On these 13 motions, the RMS error of the active finger joints on $\mathcal{T}_{\mathrm{prov}}$ (Equation~\ref{eq:provenance_subset}) is 0.059\,rad with score weighting and 0.066\,rad with rejection.

\FloatBarrier
\section{Qualitative results}
\label{app:qualitative}

\FloatBarrier
\subsection{Geometry-repair cases}
\label{app:geometry_cases}

\begin{figure}[!ht]
\centering
\includegraphics[width=\linewidth]{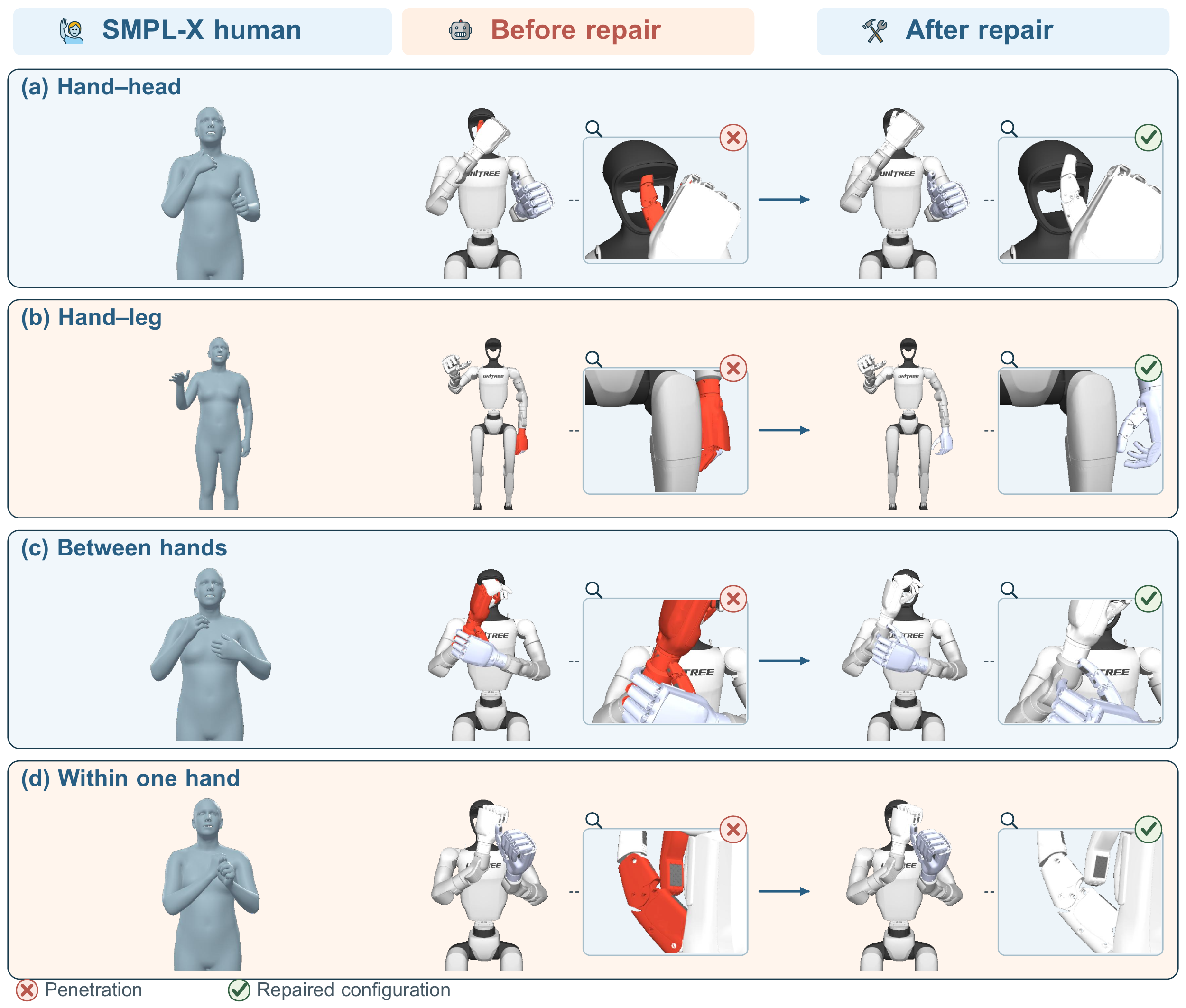}
\caption{Human fits and robot poses before and after geometry repair. Red links and crosses mark penetration; checks indicate its removal.}
\label{fig:penetration_cases}
\end{figure}

Figure~\ref{fig:penetration_cases} shows the four cases of Figure~\ref{fig:geometry_repair_main} with their source human fits and full-body robot views: hand-head, hand-leg, inter-hand, and intra-hand penetration. Human fits use the nearest source time, and paired robot views share reference time, camera, and lighting. Corpus-level statistics are in Appendix~\ref{app:geometry}.

\FloatBarrier
\subsection{Hardware replay}
\label{app:real}
\label{app:real_protocol}

\begin{figure}[!ht]
\centering
\includegraphics[width=\linewidth]{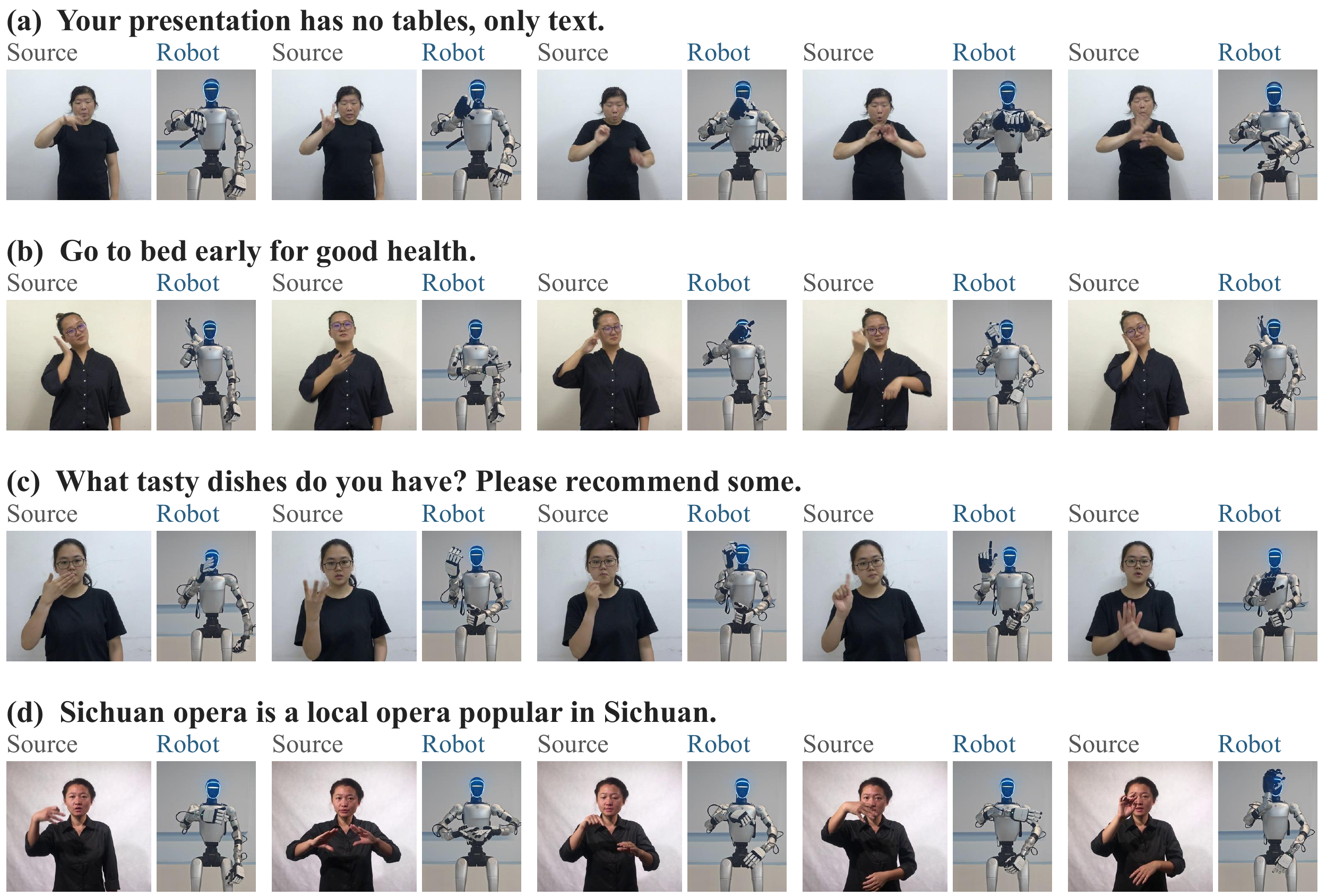}
\caption{Source signing and hardware replay for four utterances. Each row contains five chronological frame pairs, with the source signer on the left and the real robot on the right. Row labels summarize the annotated utterance meanings. Pairs use the global motion-phase alignment described above; fixed crops retain the upper body and hands while excluding wall lettering.}
\label{fig:real_robot_sequences}
\end{figure}

We recorded seven hardware replays of six motions from $\mathcal{D}_{\mathrm{control}}$ on a Unitree G1 with two Inspire RH56DFTP hands; Figure~\ref{fig:real_robot_sequences} shows four of them. Each motion is driven onboard by a per-motion tracking policy trained as in Appendix~\ref{app:control}. For replay, the reference is extended with a standing hold and an entry transition, the signing interval is played at about half speed, and an exit transition and a final hold follow. The hardware replays are qualitative and are not scored.

For the figure and the videos, the source video is aligned to each recording by one global time scale and offset, fitted between the upper-body motion in the recording and the wrist and elbow speed of the reference. No local warping is applied, the robot recording keeps its original speed, and the first and last source frames are held during the added preparation and return phases.

\FloatBarrier
\subsection{Human-motion repair casebook}
\label{app:human_casebook}

\begin{figure}[p]
\centering
\includegraphics[width=\linewidth]{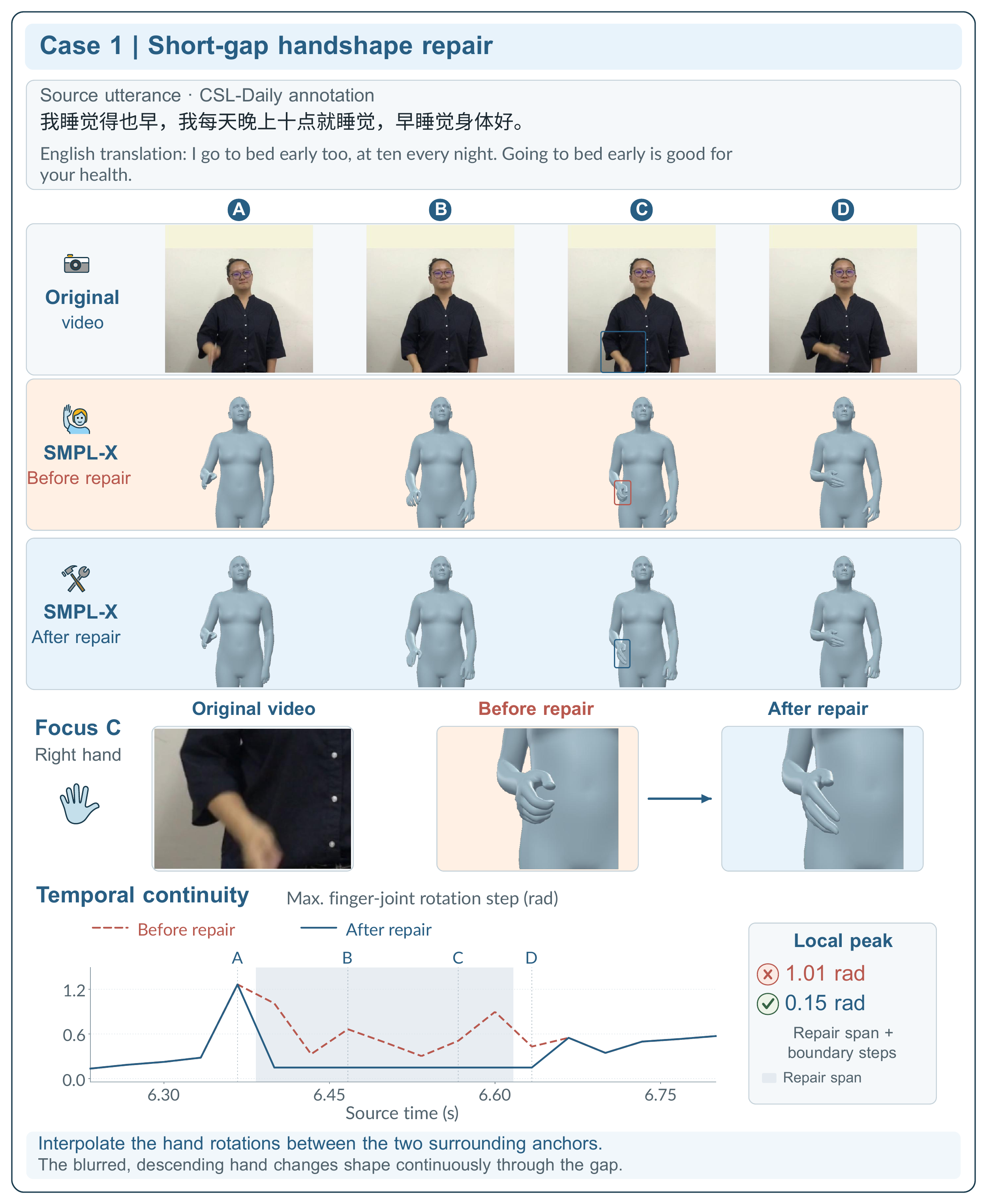}
\caption{Short-gap interpolation reduces the local hand-rotation step peak from 1.01 to 0.15 rad.}
\label{fig:human_short_gap}
\end{figure}

\begin{figure}[p]
\centering
\includegraphics[width=\linewidth]{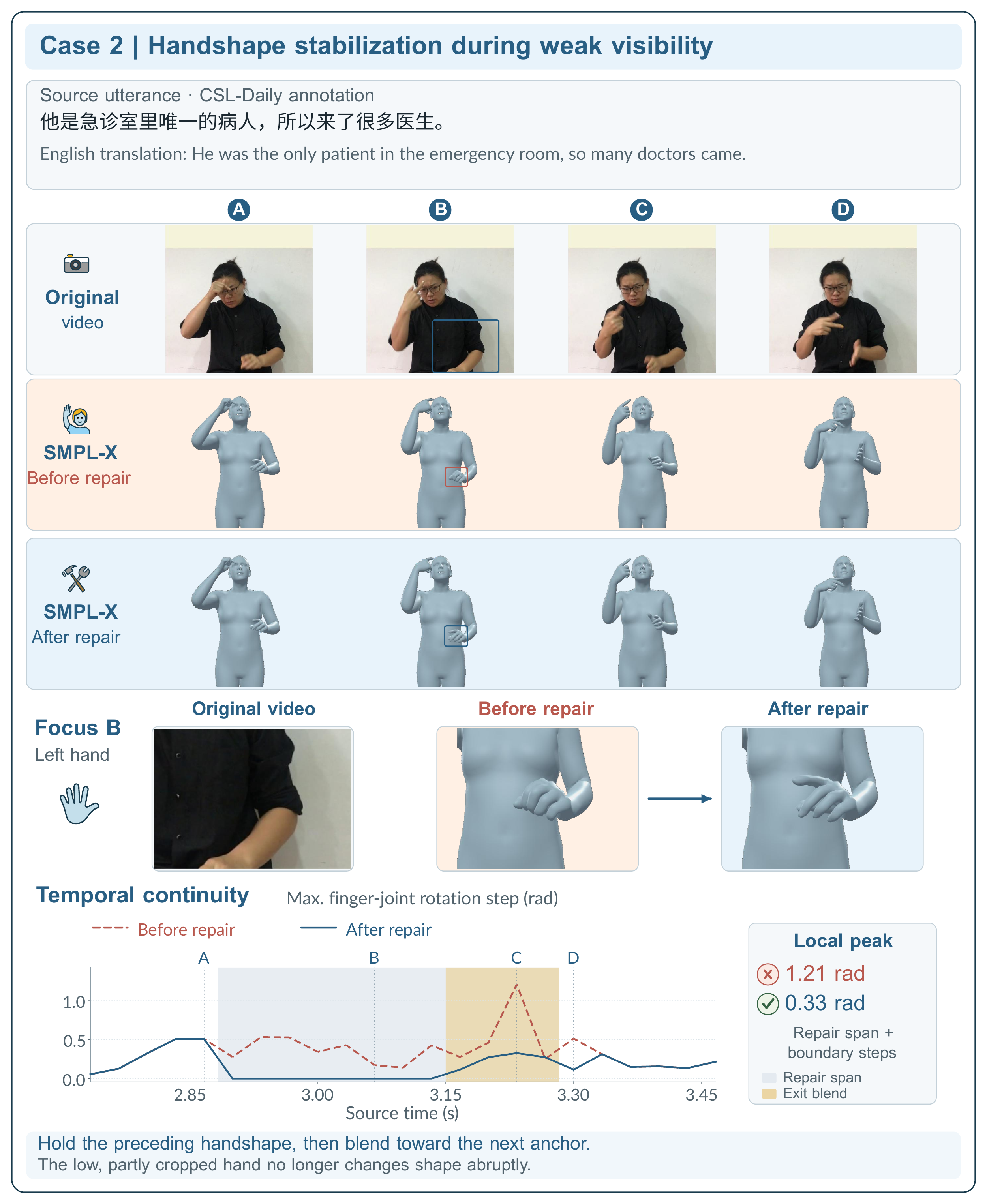}
\caption{Holding and exit blending stabilize a partly out-of-frame hand, reducing the local rotation-step peak from 1.21 to 0.33 rad.}
\label{fig:human_hand_hold}
\end{figure}

\begin{figure}[p]
\centering
\includegraphics[width=\linewidth]{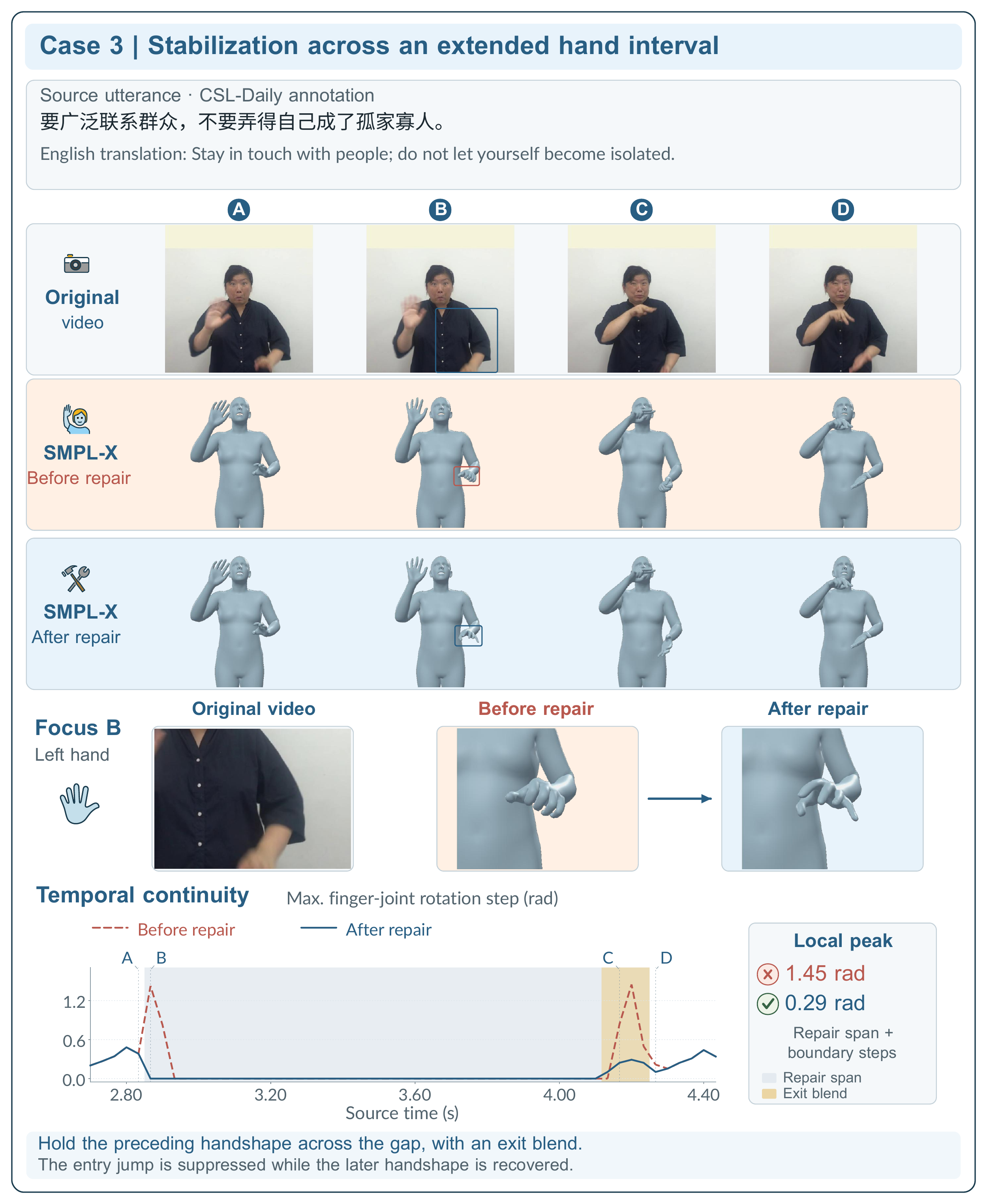}
\caption{Handshape holding and exit blending bridge an extended interval, reducing the local rotation-step peak from 1.45 to 0.29 rad.}
\label{fig:human_long_gap}
\end{figure}

\begin{figure}[p]
\centering
\includegraphics[width=\linewidth]{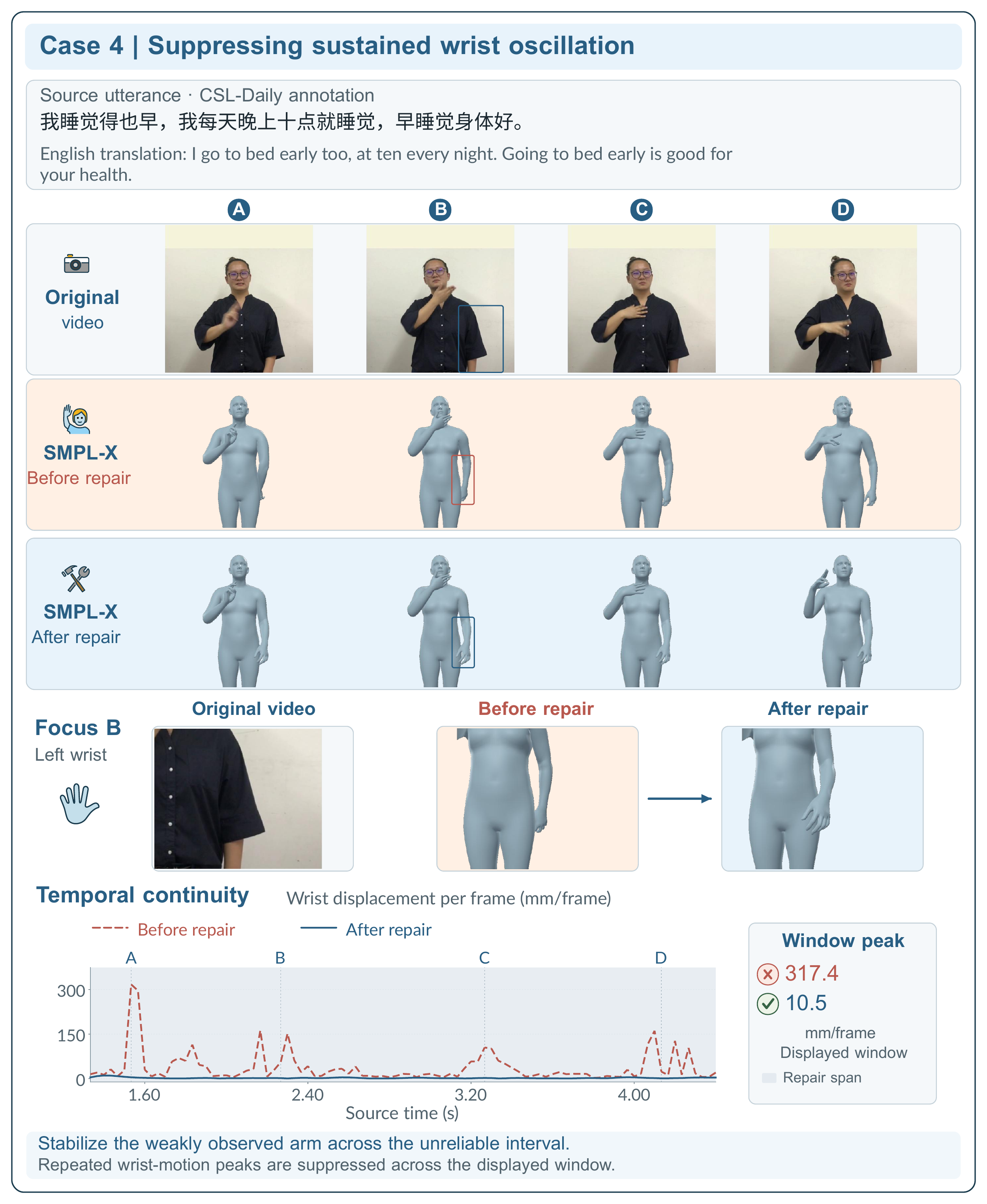}
\caption{Repair and local smoothing suppress left-wrist jitter, reducing the displayed-window displacement peak from 317.4 to 10.5\,mm/frame.}
\label{fig:human_wrist_oscillation}
\end{figure}

\begin{figure}[p]
\centering
\includegraphics[width=\linewidth]{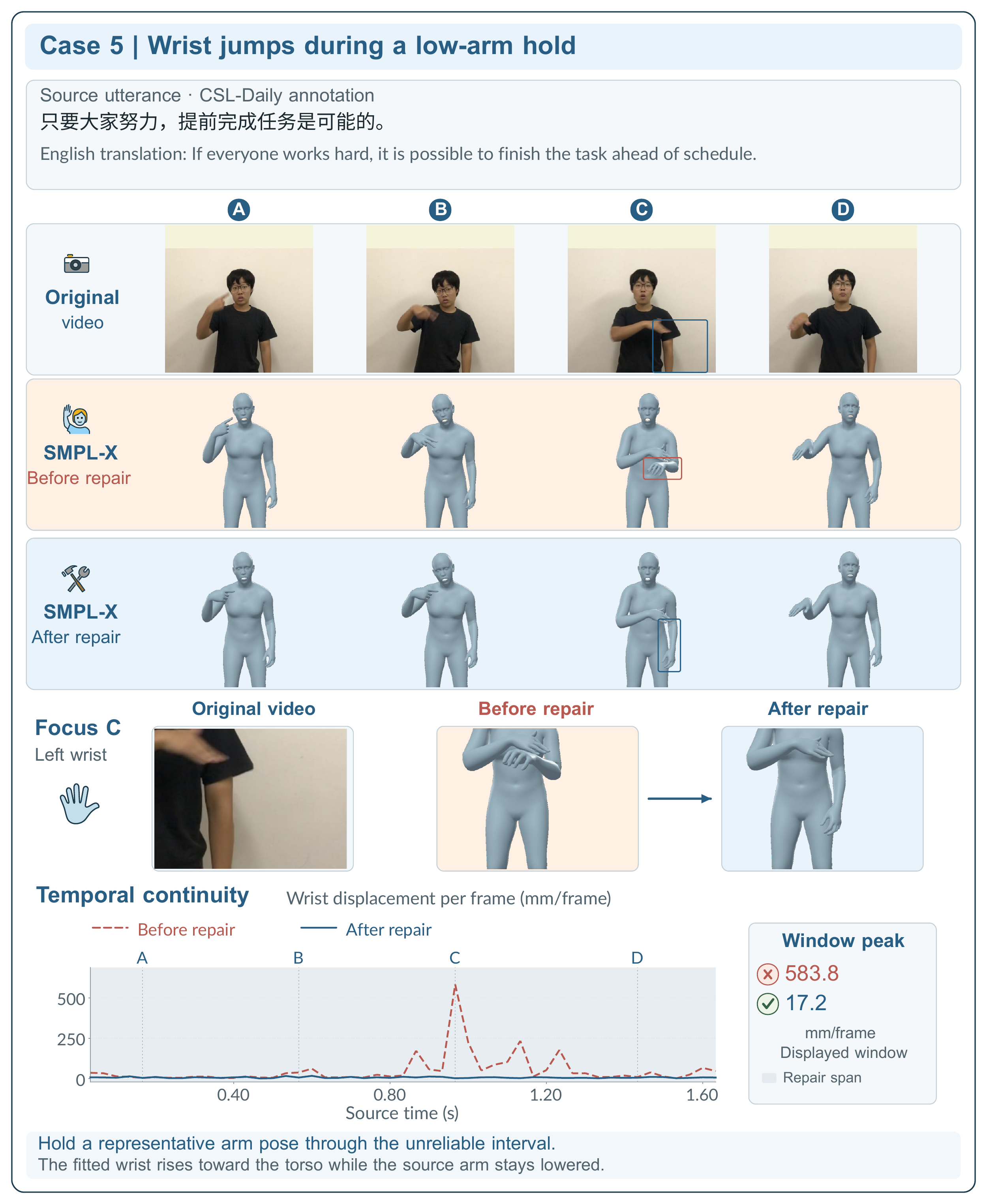}
\caption{Holding a representative low-arm pose suppresses wrist excursions toward the torso, reducing the displacement peak from 583.8 to 17.2 mm/frame.}
\label{fig:human_low_arm_jumps}
\end{figure}

\begin{figure}[p]
\centering
\includegraphics[width=\linewidth]{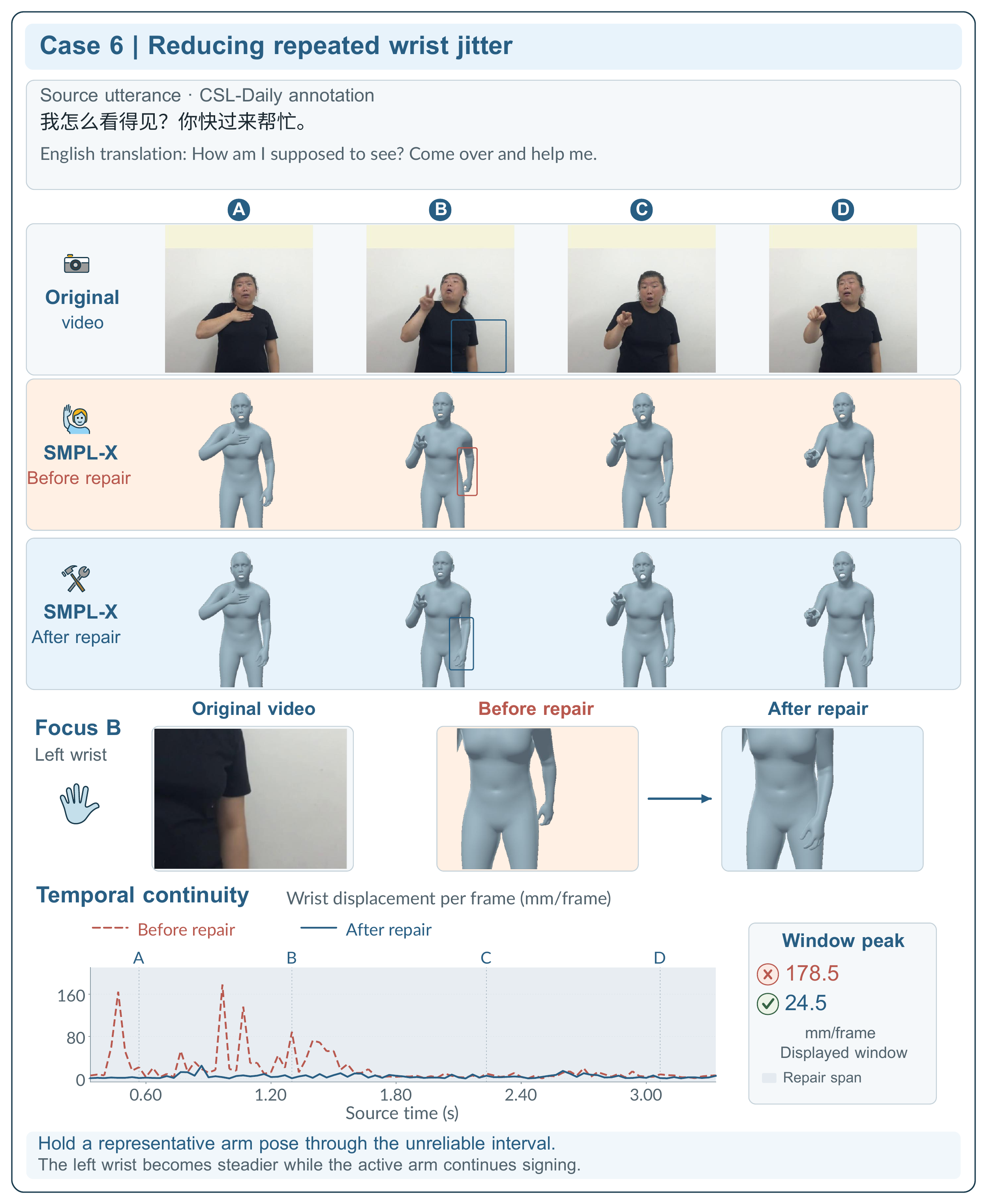}
\caption{A left-arm hold suppresses wrist jitter while the right arm signs, reducing the displacement peak from 178.5 to 24.5 mm/frame.}
\label{fig:human_repeated_wrist_jitter}
\end{figure}

\begin{figure}[p]
\centering
\includegraphics[width=\linewidth]{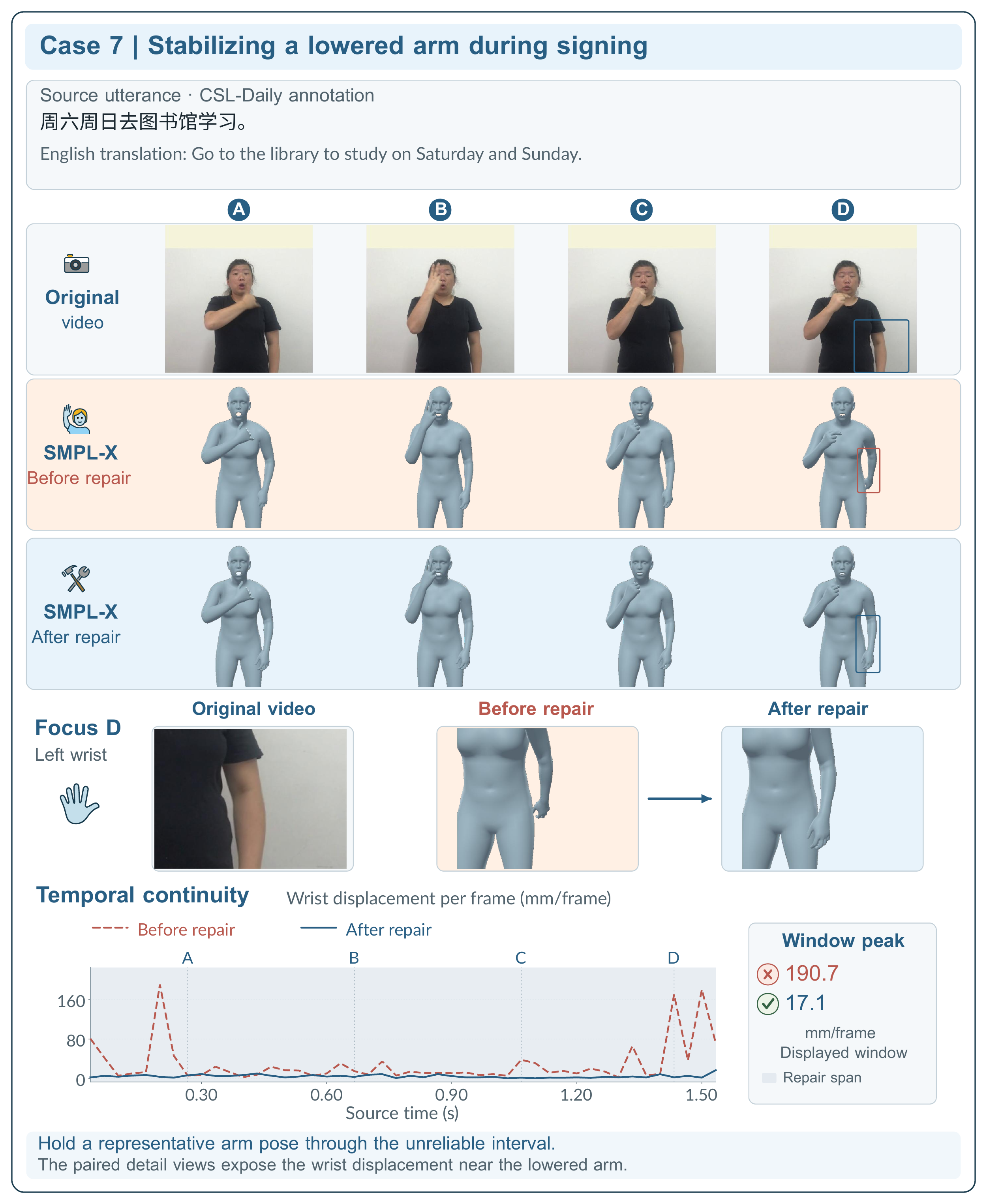}
\caption{Holding a lowered arm reduces peak wrist displacement from 190.7 to 17.1~mm/frame.}
\label{fig:human_lowered_arm}
\end{figure}

Figures~\ref{fig:human_short_gap}--\ref{fig:human_lowered_arm} show seven repaired intervals from six source sentences; Figures~\ref{fig:human_short_gap} and~\ref{fig:human_wrist_oscillation} come from the same sentence. Each plate aligns source video frames with the SMPL-X fits before and after repair, enlarges the affected part, and plots a per-frame diagnostic at 30\,Hz. Letters A--D mark matched instants, and paired meshes share camera and lighting. Gray shading marks repair intervals and sand shading marks exit blends. Figures~\ref{fig:human_short_gap}--\ref{fig:human_long_gap} plot the largest hand-joint rotation step, including the steps into and out of the repaired span. Figures~\ref{fig:human_wrist_oscillation}--\ref{fig:human_lowered_arm} plot the pelvis-relative wrist displacement per frame, with peaks taken over the displayed window. The Chinese sentences are the official CSL-Daily annotations of the complete utterances, with English translations.

The plates illustrate the repair operations and were produced with earlier settings of the same repair: Figures~\ref{fig:human_short_gap}--\ref{fig:human_long_gap} repair only the hand groups, and Figures~\ref{fig:human_wrist_oscillation}--\ref{fig:human_lowered_arm} repair the arm and hand groups with the same two cues, with light local smoothing added in Figure~\ref{fig:human_wrist_oscillation}. The released repair uses the settings of Appendix~\ref{app:human}.

\end{document}